\documentclass[11pt]{article}

\usepackage[final]{acl}

\usepackage{times}
\usepackage{latexsym}

\usepackage[T1]{fontenc}

\usepackage[utf8]{inputenc}

\usepackage{microtype}

\usepackage{inconsolata}

\usepackage{graphicx}
\usepackage{amsmath}
\usepackage{amssymb}
\usepackage{multirow}
\usepackage{booktabs}
\usepackage{algorithm}
\usepackage{algpseudocode}
\usepackage{listings}
\usepackage{xcolor}
\usepackage[table]{xcolor}
\usepackage{tcolorbox}
\definecolor{MyDarkOrange}{RGB}{166, 82, 0}

\newcommand{\method}{MetaRAG}
\newcommand{\search}{\textsc{Search}}
\newcommand{\answer}{\textsc{Answer}}
\newcommand{\best}[1]{\textbf{#1}}
\newcommand{\second}[1]{\underline{#1}}

\title{MetaRAG: Belief-Action Aligned Policy Optimization for Agentic RAG}

\author{
    Qiuyi Qi$^\spadesuit$$^\diamondsuit$\footnotemark[1],
    Tian Liang$^\spadesuit$\thanks{Q. Qi, T. Liang and J. Wang contributed equally to this work.},
    Jiamu Wang$^\spadesuit$$^\diamondsuit$\footnotemark[1], 
    Jinjian Zhang$^\diamondsuit$,
    Wei Zhou$^\diamondsuit$, \\
    \textbf{Pengcheng Zhu}$^\diamondsuit$,
    \textbf{Linjian Mo}$^\diamondsuit$,
    \textbf{Ming Kong}$^\spadesuit$\footnotemark[2],
    \textbf{Jie Liu}$^\clubsuit$\footnotemark[2], 
    \textbf{Qiang Zhu}$^\spadesuit$\thanks{Q. Zhu, M. Kong and J. Liu are corresponding authors. Q. Zhu is with the College of Artificial Intelligence, Shanghai Institute for Advanced Study, Zhejiang University. M. Kong is with the School of Earth Sciences, Zhejiang University. J. Liu is with the Department of Computer Science, City University of Hong Kong.}\\
    $^\spadesuit$ Zhejiang University, 
    $^\diamondsuit$ Ant Group,
    $^\clubsuit$ City University of Hong Kong \\
    \texttt{
    \{qiqiuyi,zhuq\}@zju.edu.cn 
    }
}

\begin{document}
\maketitle

\begin{abstract}
Agentic retrieval-augmented generation (RAG) requires language models to decide when to continue searching and when to answer. Existing RL-based methods rely on external supervision and overlook the agent's internal belief about whether the current evidence is sufficient. To address this problem, we reformulate the search decision quality as \emph{belief-action alignment} and propose \method{}, a belief-action aligned policy optimization framework for agentic RAG. \method{} uses Verify-first Action Generation to elicit an explicit verification process before each actual action, and Internal Belief Probing to estimate the policy model's own answerability belief from the same question-history context. Based on these, \method{} derives a consistency reward that is further gated by answer correctness, avoiding reinforcement of internally consistent but incorrect trajectories. The belief probe is used only during training and introduces no inference-time overhead. Experiments on seven public QA benchmarks show that \method{} consistently improves the accuracy--efficiency trade-off over strong RL-based agentic RAG baselines, with gains that transfer to deep research settings, different optimizers, and multiple model backbones.
\end{abstract}

\begin{figure}[t]
\centering
  \includegraphics[width=\columnwidth]{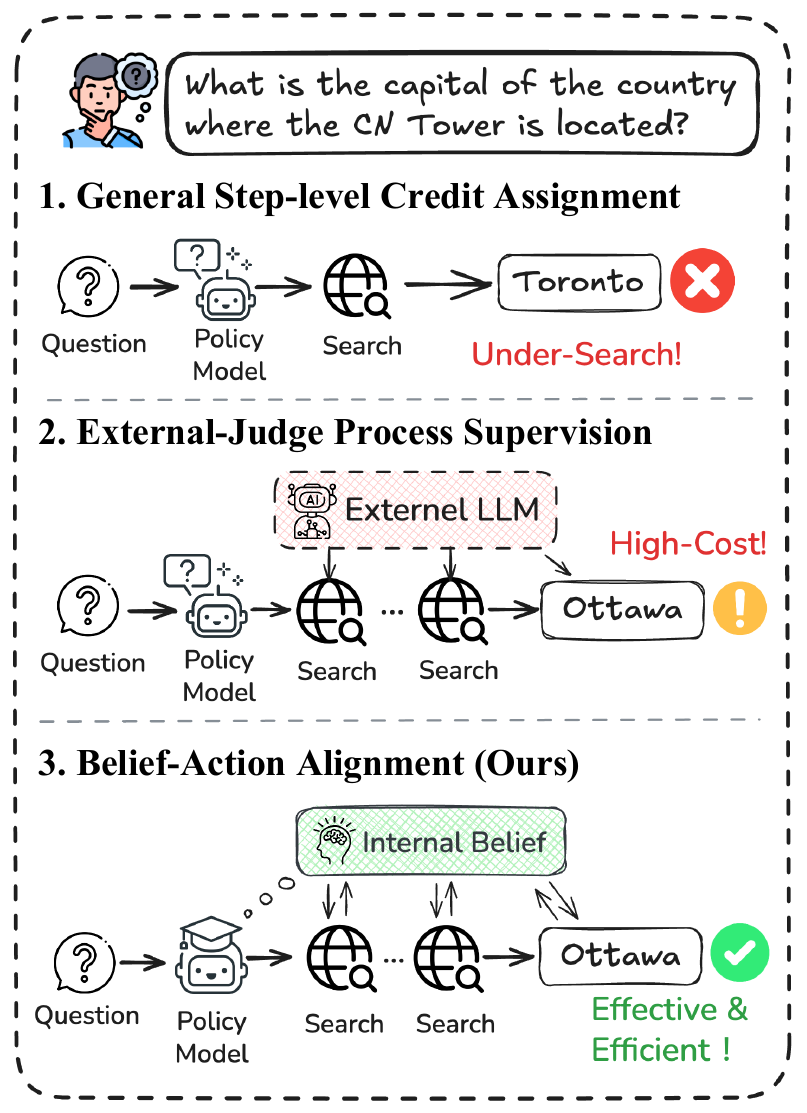}
  \caption{\textbf{Comparison of three supervision paradigms in RL-based agentic RAG.} \method{} provides lightweight decision-level supervision by aligning each Search/Answer action with the policy model's own belief about evidence sufficiency, improving boundary calibration.}
  \label{fig:motivation}
\end{figure}

\section{Introduction}

Retrieval augmented generation (RAG) grounds language model outputs in external evidence and has become a standard paradigm for knowledge-intensive question answering~\citep{lewis2020retrieval}. For complex questions, a model must operate as a search agent, which issues search queries, reads retrieved evidence, and repeatedly decides whether to continue searching or produce an answer~\citep{yao2022react,trivedi2023interleaving,asai2024self,li2025search}. Each such decision trades off evidence sufficiency against retrieval cost, making the \search{}/\answer{} decision boundary a central problem in agentic RAG.

A key challenge is boundary miscalibration. If the agent keeps retrieving when the current context is already sufficient, it incurs unnecessary cost and may introduce distracting evidence, termed as \emph{over-search}~\citep{wu2025search,qian2025smart}. If it answers before the evidence is sufficient, it risks producing unreliable outputs, called \emph{under-search}~\citep{wu2025search,shen2024smartcal}. Existing RL-based methods address this issue from different angles, as shown in Figure \ref{fig:motivation}. General step-level credit assignment methods estimate fine-grained advantages from trajectory returns or repeated states, which can suppress redundant retrieval but do not explicitly target premature answering~\citep{guan2025deeprag,feng2025group}. External-judge process supervision methods diagnose search behavior more directly with LLM judges or causal intervention, but rely on costly diagnostic procedures~\citep{wu2025hiprag,zhang2026search}. 

Across these approaches, the \search{}/\answer{} decision is supervised from the outside, while the agent's internal belief about evidence sufficiency, i.e., the signal that should drive this decision, is left unused. As a result, the agent's actions can drift away from its own beliefs, and such belief-action mismatches remain insufficiently addressed.

To address this gap, we propose to frame search decision quality as \emph{belief-action alignment}. At each decision step, the agent's action should match its own answerability belief, choosing \answer{} when it believes the current context is sufficient and \search{} when it believes the evidence is still insufficient. Under this view, over-search and under-search are two directions of the same mismatch between internal belief and actual action.

Building on this perspective, we introduce \method{}, a belief-action aligned training framework for agentic RAG. \method{} first uses \emph{Verify-first Action Generation}, where the policy model verifies a proposed candidate action before committing to an actual \search{} or \answer{} decision. It then performs \emph{Internal Belief Probing}, an independent yes/no forward pass over the same question and history, to obtain a Belief Score $P(\mathrm{Yes})-P(\mathrm{No})$. During training, this score is compared with the actual action to assign consistency credit, which is further gated by answer correctness to avoid rewarding internally consistent but incorrect trajectories. This design obtains its diagnostic signal from the policy model itself, covers both over-search and under-search, and requires no additional belief-probing pass at inference time. 

Experiments on seven public QA benchmarks show that \method{} consistently improves the accuracy--efficiency trade-off over strong RL-based agentic RAG baselines. Further analyses suggest that these gains are associated with better belief-action alignment: \method{} mitigates under-search during training, benefits from both verify-first reasoning and correctness-gated consistency rewards, strengthens knowledge-boundary awareness, and generalizes to harder deep research settings as well as different optimizers and model backbones.

Our contributions are summarized as follows:
\begin{itemize}
    \item We formulate search decision quality in agentic RAG as \emph{belief-action alignment}, unifying over-search and under-search as two forms of \search{}/\answer{} boundary mismatch.
    \item We propose \method{}, which consists of Verify-first Action Generation, Internal Belief Probing, and a correctness-gated consistency reward for scalable agentic RAG training without external judges.
    \item Extensive experiments demonstrate that \method{} improves answer accuracy, mitigates under-search, strengthens knowledge-boundary awareness, and generalizes across benchmarks, optimizers, and model backbones.
\end{itemize}

\begin{figure*}[t]
\centering
\includegraphics[width=\textwidth]{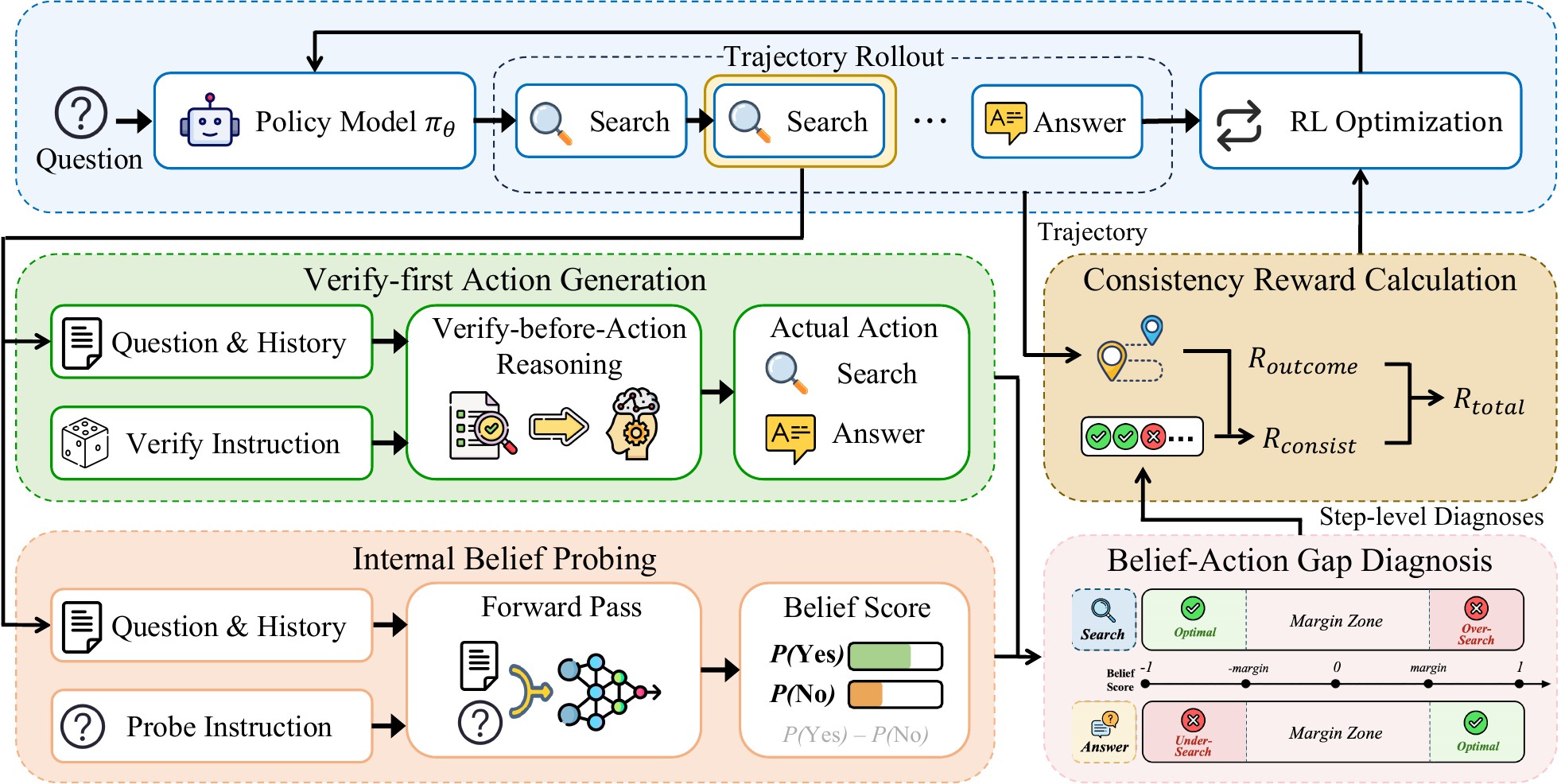}
\caption{\textbf{Overview of \method{}}. At each decision step, the policy model produces an Actual Action through Verify-first Action Generation, while Internal Belief Probing estimates a Belief Score from a separate yes/no Forward Pass. Belief-Action Gap Diagnosis maps the relation between the Belief Score and the Actual Action into a Step-level Diagnosis, and Consistency Reward Calculation combines $R_{{outcome}}$ with $R_{{consist}}$ into $R_{{total}}$ for training.}
\label{fig:framework}
\end{figure*}

\section{Related Work}
\paragraph{Agentic RAG.}
Retrieval augmented generation grounds language model outputs with external evidence~\citep{lewis2020retrieval}, while agentic RAG further enables models to decide when and what to retrieve through multi-turn interaction. Early systems mainly rely on prompting or training-free reasoning patterns, such as interleaving reasoning and actions, decomposing questions into retrieval subgoals, or self-reflecting over retrieved evidence~\citep{yao2022react,trivedi2023interleaving,asai2024self,li2025search}. More recent work treats agentic RAG as a trainable decision-making problem. Outcome-supervised methods, such as Search-R1~\citep{jin2025search} and ZeroSearch~\citep{sun2025zerosearch}, optimize search agents with final answer rewards, but provide sparse guidance for intermediate Search/Answer decisions. Process-supervised methods introduce denser signals: HiPRAG~\citep{wu2025hiprag} and DAS~\citep{zhang2026search} diagnose over-search and under-search using external LLM judges or causal intervention, while GiGPO~\citep{feng2025group}, StepSearch~\citep{wang2025stepsearch}, and TREEPS-RAG~\citep{zhang2026treeps} assign finer-grained credit from repeated states, subquestions, or rollout trees. \method{} is closest to process-supervised agentic RAG training, but differs by deriving a lightweight decision-level diagnostic from the policy model's own answerability belief, rather than external judges or counterfactual trajectories.

\paragraph{Adaptive Retrieval and Knowledge Boundaries.}
A related line of work studies whether LLMs can assess their own knowledge boundaries and adapt retrieval accordingly. Prior methods use uncertainty, self-reflection, metacognitive regulation, or query complexity to trigger retrieval, critique evidence, revise responses, or route questions to different retrieval pipelines~\citep{jiang2023active,asai2024self,zhou2024metacognitive,su2024dragin,jeong2024adaptive}. These signals have also been studied for adaptive inference, model cascading, and abstention~\citep{kadavath2022language,lin2022teaching,kuhn2023semantic,manakul2023selfcheckgpt,dohan2022language,gupta2024language,wen2025know,chen2025query}. Search Wisely~\citep{wu2025search} is especially close to our setting: it proposes $\beta$-GRPO, which rewards correct trajectories whose generated search queries exceed a confidence threshold. However, this line of work mainly uses such signals for inference-time control, output revision, query routing, or search-query confidence scoring. In agentic RAG, the boundary is dynamic: after each retrieval step, the agent must decide whether the current question-history context is sufficient for answering. \method{} follows this decision-boundary perspective~\citep{zhang2026search,wu2025search,wu2025hiprag}, but operationalizes it as a training-time belief-action alignment signal: the policy model's answerability belief is probed at each step and compared with its actual Search/Answer action to diagnose both over-search and under-search.

\section{Methodology}
\label{sec:method}

In this section, we present \method{}, a belief-action aligned training framework for agentic RAG. \method{} calibrates the \search{}/\answer{} boundary by coupling two complementary signals: Verify-first Action Generation (\S\ref{sec:verify-first}) elicits an Actual Action after Verify-before-Action Reasoning, while Internal Belief Probing (\S\ref{sec:internal_belief}) estimates whether the same model believes the current Question \& History are answerable. Their mismatch is diagnosed as a belief-action gap (\S\ref{sec:gap_diag}) and converted into a consistency reward (\S\ref{sec:consist_reward}) for policy optimization. Figure~\ref{fig:framework} illustrates the overall workflow.

\subsection{Task Definitions}

Given a question $q$, an agentic RAG policy model $\pi_\theta$ interacts with a retriever for at most $T$ turns. At turn $t$, the decision context is $c_t=(q,h_t)$, where $h_t$ contains previous search history. The model emits an \emph{Actual Action} $a_t$ with action type $\alpha_t\in\{\search,\answer\}$. If $\alpha_t=\search$, then $a_t=\search(s_t)$ and the retriever returns the top-$K$ passages for query $s_t$. If $\alpha_t=\answer$, then $a_t=\answer(y_t)$ and the episode terminates. A trajectory is denoted as $\tau=(c_1,a_1,\ldots,c_{T_\tau},a_{T_\tau})$, where $T_\tau\le T$.

\subsection{Verify-first Action Generation}
\label{sec:verify-first}

At each decision step, \method{} first proposes a candidate action type $\tilde{\alpha}_t\in\{\search,\answer\}$ and then asks the policy model to perform \emph{Verify-before-Action Reasoning} before committing to its Actual Action. The input consists of the Question \& History and the following Verify Instruction:
\begin{tcolorbox}[colback=gray!5!white, colframe=black!75!black, 
title=Verify Instruction, boxrule=0.3mm, width=\columnwidth, arc=3mm, auto outer arc=true]
Before making your own decision for this step, a candidate action is proposed as: \{candidate\_decision\}. You should first conduct reasoning, starting by verifying whether the proposed candidate action is appropriate for the current step, then determine the correct action.
\end{tcolorbox}
The candidate action is a decision hypothesis rather than a forced label. The model may accept or reject it after verification, and the generated response contains both a verification rationale and the Actual Action $a_t$. We use this strategy during both training and inference to make evidence sufficiency explicit before each Search/Answer decision.

\subsection{Internal Belief Probing}
\label{sec:internal_belief}
Internal Belief Probing estimates the model's answerability belief independently of action generation. For the same context $c_t=(q,h_t)$, we run one additional Forward Pass with the Question \& History and the following Probe Instruction:
\begin{tcolorbox}[colback=gray!5!white, colframe=black!75!black, 
title=Probe Instruction, boxrule=0.3mm, width=\columnwidth, arc=3mm, auto outer arc=true]
Respond ONLY with `Yes' or `No' to indicate whether you are capable of answering the question confidently.
\end{tcolorbox}

Let $\ell_t^Y$ and $\ell_t^N$ be the next-token logits of ``Yes'' and ``No''. We normalize these two logits as
\begin{equation}
    P_t(\mathrm{Yes}), P_t(\mathrm{No})
    =\mathrm{softmax}(\ell_t^Y,\ell_t^N),
\end{equation}
and define the Belief Score as
\begin{equation}
    b_t=P_t(\mathrm{Yes})-P_t(\mathrm{No})\in[-1,1].
\end{equation}
A larger $b_t$ indicates that the model believes the current context is more answerable. Since the probe uses the same policy model rather than an external verifier, it reflects the model's own knowledge boundary. The belief probe is used only for training-time reward calculation; inference requires no additional probing Forward Pass. Appendix~\ref{sec:appendix_alt_probe} further evaluates an Internal Confidence probe~\citep{chen2025query}, showing that \method{} is not tied to this particular yes/no logit implementation.

\subsection{Belief-Action Gap Diagnosis}
\label{sec:gap_diag}
Belief-Action Gap Diagnosis compares the Belief Score with the Actual Action.
Let $m\in[0,1)$ be a margin. The regions $b_t>m$ and $b_t<-m$ indicate confident answerability and confident insufficiency, respectively, while $|b_t|\le m$ defines a Margin Zone where the belief signal is treated as inconclusive.

We define a binary consistency indicator $o_t$ to assign positive consistency credit only when the Actual Action is aligned with a confident belief signal:
\begin{equation}
    o_t =
    \begin{cases}
    1, & b_t>m,\ \alpha_t=\answer,\\
    1, & b_t<-m,\ \alpha_t=\search,\\
    0, & \text{otherwise}.
    \end{cases}
\end{equation}
When $b_t>m$ but $\alpha_t=\search$, the step is diagnosed as \textsc{Over-search}, since the model retrieves despite believing the current context is answerable. When $b_t<-m$ but $\alpha_t=\answer$, the step is diagnosed as \textsc{Under-search}, since the model answers despite believing more evidence is needed. Steps with $o_t=1$ are diagnosed as \textsc{Optimal}. Qualitative examples of both diagnoses are provided in Appendix~\ref{sec:appendix_case_bag_diagnosis}.

\begin{table*}[t]
\centering

\resizebox{\textwidth}{!}{
\begin{tabular}{l*{3}{cc}|*{4}{cc}|cc}
\toprule
\multirow{3}{*}{\textbf{Method}}
& \multicolumn{6}{c|}{\textbf{Single-Hop QA}}
& \multicolumn{8}{c|}{\textbf{Multi-Hop QA}}
& \multicolumn{2}{c}{\multirow{2}{*}{\textbf{Avg.}}} \\
\cmidrule(lr){2-7}\cmidrule(lr){8-15}
& \multicolumn{2}{c}{NQ$^{\dagger}$}
& \multicolumn{2}{c}{TriviaQA$^{\star}$}
& \multicolumn{2}{c|}{PopQA$^{\star}$}
& \multicolumn{2}{c}{HotpotQA$^{\dagger}$}
& \multicolumn{2}{c}{2Wiki$^{\star}$}
& \multicolumn{2}{c}{MuSiQue$^{\star}$}
& \multicolumn{2}{c|}{Bamboogle$^{\star}$}
& \multicolumn{2}{c}{} \\
\cmidrule(lr){2-3}\cmidrule(lr){4-5}\cmidrule(lr){6-7}
\cmidrule(lr){8-9}\cmidrule(lr){10-11}\cmidrule(lr){12-13}
\cmidrule(lr){14-15}\cmidrule(lr){16-17}
& Acc. & Searches
& Acc. & Searches
& Acc. & Searches
& Acc. & Searches
& Acc. & Searches
& Acc. & Searches
& Acc. & Searches
& Acc. & Searches \\
\midrule

\multicolumn{17}{l}{\textit{\textbf{Qwen2.5-3B-Instruct}}} \\
Search-R1
& 39.7 & 1.55
& 56.5 & 1.53
& 39.1 & 1.75
& 33.1 & 2.78
& 31.0 & 3.38
& 12.4 & 3.39
& 23.2 & 2.79
& 33.6 & 2.45 \\

HiPRAG
& 43.0 & 1.01
& 59.8 & 1.01
& 42.0 & 1.02
& 36.0 & 2.14
& \best{40.5} & 2.48
& 10.8 & 2.55
& 24.0 & 2.02
& 36.6 & 1.75 \\

GiGPO$^{\ddagger}$
& \second{45.1} & 1.00
& \second{60.9} & 0.99
& \second{44.9} & 1.17
& \second{38.0} & 1.24
& 37.1 & 1.55
& \second{14.5} & 1.81
& \second{37.9} & 1.49
& \second{39.8} & 1.32 \\

\method{}
& \best{45.6} & 1.17
& \best{62.8} & 1.19
& \best{48.0} & 1.27
& \best{41.3} & 1.58
& \second{39.1} & 1.87
& \best{15.9} & 2.24
& \best{39.5} & 1.90
& \best{41.7} & 1.60 \\

\midrule

\multicolumn{17}{l}{\textit{\textbf{Qwen2.5-7B-Instruct}}} \\
Search-R1
& 42.9 & 1.70
& 62.3 & 1.65
& 42.7 & 1.90
& 38.6 & 2.95
& 34.6 & 3.55
& 16.2 & 3.60
& 40.0 & 2.95
& 39.6 & 2.61 \\

HiPRAG
& 46.5 & 1.80
& \second{65.8} & 1.79
& 45.8 & 1.88
& 42.0 & 2.28
& \best{46.1} & 2.54
& 14.0 & 2.59
& 40.0 & 2.30
& 42.9 & 2.17 \\

GiGPO$^{\ddagger}$
& \second{46.8} & 0.96
& \second{65.8} & 0.71
& \second{48.1} & 1.14
& \second{42.3} & 1.20
& 42.9 & 1.40
& \second{18.2} & 2.09
& \second{43.2} & 1.22
& \second{43.9} & 1.25 \\

\method{}
& \best{46.9} & 1.62
& \best{67.2} & 1.19
& \best{48.5} & 1.85
& \best{45.7} & 1.67
& \second{44.9} & 1.69
& \best{20.3} & 2.31
& \best{44.0} & 1.83
& \best{45.4} & 1.74 \\

\bottomrule
\end{tabular}
}
\caption{\textbf{Main results on seven public QA benchmarks.} $\dagger$ and $\star$ indicate in-domain and out-of-domain datasets, respectively. Acc. denotes Exact Match (EM) accuracy (\%), and Searches denotes the average number of search calls per question. Best and second-best accuracy values are \textbf{bold} and \underline{underlined}. Results marked with $\ddagger$ are reproduced by us using the same retrieval setup; the remaining baseline results are taken from \citet{xia2026search}.}
\label{tab:main}
\end{table*}

\subsection{Consistency Reward Calculation}
\label{sec:consist_reward}
For a trajectory with $T_\tau$ identifiable decision steps, Consistency Reward Calculation aggregates Step-level Diagnoses through the binary consistency indicators $\{o_t\}_{t=1}^{T_\tau}$:
\begin{equation}
    R_{{consist}}(\tau)=\frac{1}{T_\tau}\sum_{t=1}^{T_\tau}o_t.
\end{equation}
We also compute the outcome reward $R_{{outcome}}(\tau)=\mathbb{I}[\mathrm{EM}(\hat{y},y^*)]$, where $\hat{y}$ is the final answer and $y^*$ is the gold answer. The total reward is
\begin{equation}
    R_{{total}}(\tau)
    =R_{{outcome}}(\tau) \cdot \left(1+\lambda R_{{consist}}(\tau)\right),
\end{equation}
where $\lambda$ controls the strength of belief-action alignment.

The multiplication by $R_{{outcome}}$ gates the consistency reward with final correctness: an incorrect trajectory receives zero reward even if its Search/Answer decisions appear internally consistent. This prevents reward hacking through confidently stopping early, avoiding retrieval, or following an erroneous belief without solving the task. The resulting $R_{{total}}$ is used as the scalar reward for downstream RL optimization. The full reward calculation procedure is provided in Algorithm~\ref{alg:metarag} in Appendix~\ref{sec:appendix_pseudocode}.

\section{Experiments}
\subsection{Experiment Setup}
\paragraph{Benchmarks.} 
Following prior work, we evaluate on seven public QA benchmarks spanning two categories:
(1) \textbf{Single-Hop QA}: NQ~\citep{kwiatkowski2019natural}, TriviaQA~\citep{joshi2017triviaqa}, and PopQA~\citep{mallen2022not};
(2) \textbf{Multi-Hop QA}: HotpotQA~\citep{yang2018hotpotqa}, 2Wiki~\citep{ho2020constructing}, MuSiQue~\citep{trivedi2022musique}, and Bamboogle~\citep{press2022measuring}.
Following Search-R1~\citep{jin2025search}, we train on the merged NQ and HotpotQA training sets and evaluate on all seven datasets to assess both in-domain and out-of-domain generalization.

\paragraph{Baselines.} 
We compare \method{} with three representative RL-based agentic RAG baselines.
(1) \textbf{Search-R1}~\citep{jin2025search} optimizes search-augmented QA agents with sparse outcome rewards. 
(2) \textbf{HiPRAG}~\citep{wu2025hiprag} augments outcome supervision with hierarchical process rewards, where external LLM judges are used to detect over-search and under-search behaviors during training. 
(3) \textbf{GiGPO}~\citep{feng2025group} improves multi-turn agent training through group-in-group policy optimization, enabling fine-grained credit assignment at the step level.
These baselines cover outcome-only supervision, external-judge process supervision, and general step-level credit assignment.

\paragraph{Implementation Details.}
We conduct experiments using Qwen2.5-3B-Instruct and Qwen2.5-7B-Instruct.
For retrieval, we follow Search-R1~\citep{jin2025search}, using the 2018 Wikipedia dump~\citep{karpukhin2020dense} as the knowledge source, E5~\citep{wang2022text} as the retriever, and returning the top-3 passages for each search query.
For RL training, we adopt GRPO~\citep{shao2024deepseekmath} with rollout group size $N=5$ and a maximum of $4$ turns. 
All experiments are conducted on a single node with 8 A100 GPUs. 
Unless otherwise specified, the candidate action type $\tilde{\alpha}_t$ is sampled uniformly from $\{\search,\answer\}$, the consistency reward weight $\lambda$ is set to $0.1$, and the margin $m$ is set to $0.0$.
Full training settings and hyperparameter details are provided in Appendix~\ref{appendix:ed}.

\begin{figure*}[t]
\centering
\includegraphics[width=\textwidth]{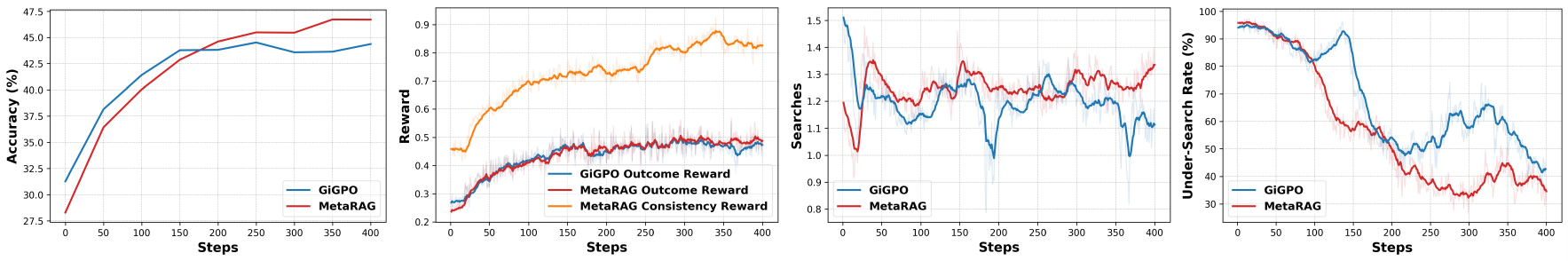}
\caption{\textbf{Training dynamics on Qwen2.5-3B-Instruct.} Solid lines denote exponential moving averages in all panels except the Accuracy panel. From left to right, the panels show accuracy, rewards, average search count, and under-search rate. \method{} initially trails GiGPO without SFT warm-up for the new verify-before-action reasoning pattern, but later surpasses it in both accuracy and outcome reward. Its consistency reward increases steadily, and its under-search rate remains mostly lower after about 100 steps.}
\label{fig:dynamics}
\end{figure*}

\begin{table}[t]
\centering

\resizebox{0.8\columnwidth}{!}{
\begin{tabular}{lcc}
\toprule
Variant & Acc. & Searches \\
\midrule
\multicolumn{3}{l}{\textit{\textbf{Core Components}}} \\
w/o Consistency Reward & 41.9 & 2.14 \\
w/o Verify-before-Action & 40.4 & 1.38 \\
\midrule
\multicolumn{3}{l}{\textit{\textbf{Diagnostic Signal}}} \\
Over-Search Only & 41.3 & 1.33 \\
Under-Search Only & 42.4 & 2.56 \\
w/ Incorrect Trajectories & 39.6 & 1.12 \\
\midrule
\multicolumn{3}{l}{\textit{\textbf{Candidate Strategy}}} \\
Always \search{} & 41.1 & 1.77 \\
Always \answer{} & 41.1 & 1.62 \\
Heuristic & 41.2 & 1.46 \\
\midrule
\textbf{w/ \method{}} & 41.7 & 1.60 \\
\bottomrule
\end{tabular}}
\caption{\textbf{Ablation study on Qwen2.5-3B-Instruct.} Results are averaged over seven QA benchmarks.}
\label{tab:ablation}
\end{table}

\subsection{Main Results}
Table~\ref{tab:main} shows that \method{} consistently delivers the strongest overall performance across both backbone sizes. On Qwen2.5-3B-Instruct and Qwen2.5-7B-Instruct, \method{} achieves average accuracies of 41.7\% and 45.4\%, outperforming the strongest baseline GiGPO by 1.9 and 1.5 points, respectively. Importantly, these gains do not come from simply increasing retrieval frequency. Compared with HiPRAG, which uses external LLM judges to supervise search behavior, \method{} achieves higher accuracy with fewer searches on both model sizes. Compared with GiGPO, \method{} performs more searches but obtains consistently better accuracy, including gains on all four multi-hop benchmarks. 
These results indicate that \method{} achieves a more favorable accuracy--efficiency trade-off among the evaluated RL-based agentic RAG methods, while the source of this improvement is examined in the following analyses. 
We further provide a same-setting comparison with $\beta$-GRPO~\citep{wu2025search}, an uncertainty-aware search-training method, in Appendix~\ref{sec:appendix_search_wisely}.
The framework is also compatible with GiGPO's step-level credit assignment: \method{} (GiGPO) further improves the average accuracy to 43.1\% and 47.0\% on Qwen2.5-3B-Instruct and Qwen2.5-7B-Instruct, respectively (Appendix~\ref{sec:appendix_gigpo}).

\subsection{Training Dynamics}
Figure~\ref{fig:dynamics} shows the training dynamics on Qwen2.5-3B-Instruct. 
\method{} starts below GiGPO in accuracy, likely because Verify-first Action Generation introduces a new reasoning pattern without SFT warm-up. 
After the policy adapts, \method{} overtakes GiGPO around the middle of training and converges to higher accuracy. 
The outcome reward largely mirrors this trend: GiGPO is higher in the first $\sim$175 steps, while \method{} becomes higher from roughly 175 to 400 steps. 
Meanwhile, the consistency reward steadily increases, suggesting that the belief-action signal provides a stable auxiliary training signal. 
Although \method{} generally uses a moderately larger search budget than GiGPO, its under-search rate falls below GiGPO after about 100 steps and remains mostly lower thereafter. 
Here, under-search rate is a training-time belief-based diagnostic, measuring answer steps where the model chooses \answer{} while its Belief Score indicates insufficient answerability ($b_t<-m$). 
To avoid relying only on this internal signal, we further conduct an external LLM-as-judge validation on test trajectories in Appendix~\ref{sec:appendix_external_undersearch}; the same trend holds under this external assessment. 
Together with the qualitative comparison in Appendix~\ref{sec:appendix_case_under_search_mitigation}, these results indicate that the additional retrieval is associated with fewer premature answer actions, consistent with the stronger accuracy--efficiency trade-off observed in Table~\ref{tab:main}.

\subsection{Ablation Study}
We validate the main design choices of \method{} on Qwen2.5-3B-Instruct, with all results averaged across seven QA benchmarks. Detailed per-dataset results are provided in Appendix~\ref{sec:appendix_dasr}.

\paragraph{Core Components.}
The top block of Table~\ref{tab:ablation} shows that both the consistency reward and verify-before-action are necessary for a favorable accuracy--efficiency trade-off. Removing the consistency reward slightly increases accuracy but raises the average number of searches from 1.60 to 2.14, indicating that verify-first reasoning alone tends to solve questions through more aggressive retrieval. Removing verify-before-action reduces accuracy to 40.4\%, showing that the verification step is not merely a prompt-level addition but provides the decision context in which belief-action alignment becomes effective. We further isolate the inference-time effect of verify-first prompting in Appendix~\ref{sec:appendix_verify_inference}. Adding the same verify-first inference format to GiGPO does not improve performance, while \method{} remains strong even when verify-first reasoning is disabled at inference, suggesting that the gains are not simply due to a stronger inference prompt.

\begin{figure}[t]
\centering
  \includegraphics[width=\columnwidth]{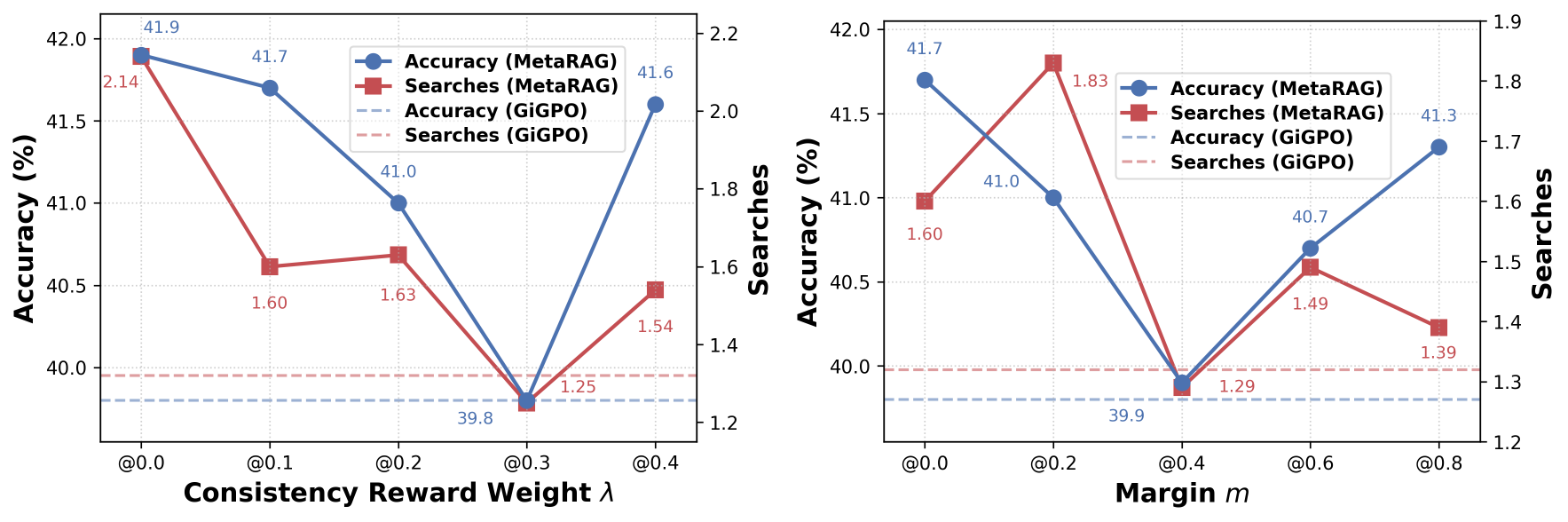}
  \caption{\textbf{Hyperparameter sensitivity on Qwen2.5-3B-Instruct.} Left: consistency reward weight $\lambda$. Right: margin $m$. Dashed lines indicate GiGPO.}
  \label{fig:hyper}
\end{figure}

\begin{figure}[t]
\centering
  \includegraphics[width=\columnwidth]{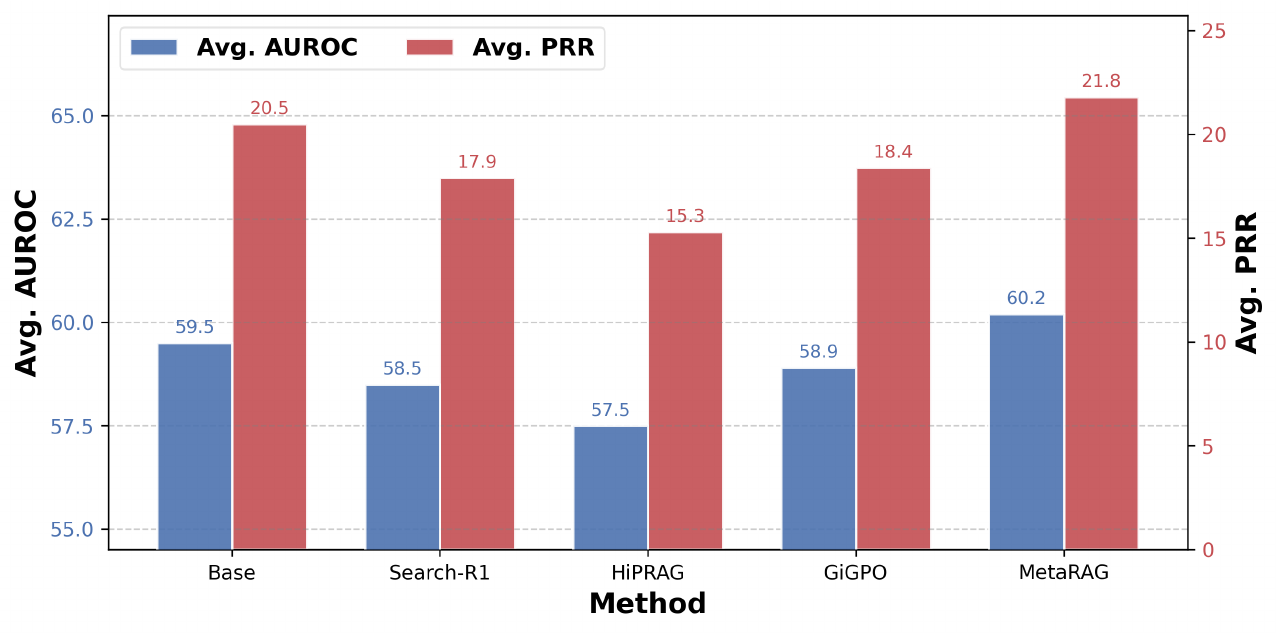}
  \caption{\textbf{Perplexity-based knowledge-boundary awareness attribution on Qwen2.5-7B-Instruct.} We report average AUROC and PRR over GSM8K, SciQ, and TriviaQA. Higher values indicate stronger separation between answerable and non-answerable queries.}
  \label{fig:kb}
\end{figure}

\paragraph{Diagnostic Signal.}
The middle block of Table~\ref{tab:ablation} examines which diagnostic signals should contribute to the consistency reward. Over-Search Only\footnote{Over-Search Only computes the consistency reward over search steps only: the numerator counts search steps not diagnosed as over-search, and the denominator is the number of search steps. Under-Search Only is defined symmetrically over answer steps.} yields a frugal but less accurate policy, suggesting that suppressing redundant retrieval alone can make the model overly conservative. Under-Search Only achieves the highest ablation accuracy, but requires substantially more searches, reflecting the opposite bias toward retrieval. Including Incorrect Trajectories is the most damaging variant, dropping accuracy to 39.6\%, which confirms the need for outcome gating: consistency credit should reinforce belief-action alignment only when the trajectory also solves the task.

\paragraph{Candidate Strategy.}
The bottom block of Table~\ref{tab:ablation} compares different ways of constructing the candidate action in Verify-first Action Generation. Always proposing \search{} or \answer{} underperforms the default random strategy. The heuristic strategy uses the previous non-terminal action as the candidate decision---typically \search{}, since an \answer{} action terminates the episode---and switches to \answer{} at the maximum turn. Although this hand-crafted continuation/termination prior reduces searches, it also lowers accuracy, suggesting that random candidates provide a more balanced verification process by exposing the model to both sides of the \search{}/\answer{} boundary.

\subsection{Hyperparameter Sensitivity}
Figure~\ref{fig:hyper} studies the consistency reward weight $\lambda$ and margin $m$. Across the tested ranges, \method{}'s accuracy is almost always higher than GiGPO, indicating that the method is not tied to a brittle hyperparameter choice. For $\lambda$, the default value $0.1$ offers a strong trade-off: it substantially reduces searches compared with $\lambda=0$ while preserving similar accuracy, whereas larger weights make the trade-off less stable. For the margin, $m=0.0$ achieves the best accuracy, but larger margins (e.g., $m=0.8$) can still perform strongly. This non-monotonic pattern suggests a coverage--reliability trade-off in belief signals: using all signed belief signals maximizes supervision, while wider Margin Zones filter uncertain signals but may discard useful boundary information. Detailed per-dataset results are provided in Appendix~\ref{sec:appendix_dhsr}.

\subsection{Attribution Analysis}
To understand why belief-action alignment improves the search--answer trade-off, we examine how RL training affects the model's knowledge-boundary awareness. Following uncertainty evaluation in \citet{chen2025query}, we use Perplexity~\citep{fomicheva2020unsupervised} as a diagnostic signal to distinguish answerable from non-answerable queries and report the Area Under the Receiver Operating Characteristic Curve (AUROC) and Prediction Rejection Ratio (PRR), averaged over GSM8K~\citep{cobbe2021training}, SciQ~\citep{welbl2017crowdsourcing}, and TriviaQA~\citep{joshi2017triviaqa}. As shown in Figure~\ref{fig:kb}, standard agentic RAG training tends to weaken this boundary signal: Search-R1, HiPRAG, and GiGPO all fall below the base model in both AUROC and PRR. In contrast, \method{} achieves the best results, improving over the base model from 59.5 to 60.2 in AUROC and from 20.5 to 21.8 in PRR. 
This suggests that \method{}'s gains are not solely due to changing search frequency; its belief-action consistency reward also helps retain and strengthen the uncertainty signal that separates when the model can answer from when retrieval is needed. 
Detailed per-dataset results, including an additional attentional-entropy diagnostic, are provided in Appendix~\ref{sec:appendix_daar}.

\begin{table}[t]
\centering
\resizebox{0.8\columnwidth}{!}{
\begin{tabular}{lccc}
\toprule
Method & Acc. (\%) & Rec. (\%) & Searches \\
\midrule
\multicolumn{4}{l}{\textit{\textbf{Qwen3-32B}}} \\
Base & \textbf{3.61} & \underline{3.12} & 0.92 \\
\midrule
\multicolumn{4}{l}{\textit{\textbf{Qwen2.5-7B-Instruct}}} \\
Search-R1 & 1.45 & 2.78 & 2.38 \\
HiPRAG & 1.69 & 2.71 & 2.62 \\
GiGPO & 2.65 & 3.10 & 2.32 \\
\method{} & \underline{3.49} & \textbf{3.73} & 3.14 \\
\bottomrule
\end{tabular}}
\caption{\textbf{Zero-shot transfer to BrowseComp-Plus with BM25 retriever.} Acc. denotes answer accuracy, Rec. denotes evidence recall, and Searches denotes the average number of search calls per question. Best and second-best Acc./Rec. values are \textbf{bold} and \underline{underlined}.}
\label{tab:browse}
\end{table}

\subsection{Zero-shot Transfer to BrowseComp-Plus}
Table~\ref{tab:browse} evaluates zero-shot transfer to BrowseComp-Plus, a challenging deep research benchmark that requires more extensive information seeking than standard QA datasets~\citep{chen2025browsecomp}. 
All Qwen2.5-7B-Instruct agents are trained only on the original QA training data and are tested with BM25 retriever. 
Among 7B search agents, \method{} achieves the best accuracy and recall, improving over GiGPO from 2.65\% to 3.49\% in accuracy and from 3.10\% to 3.73\% in recall. 
It also approaches the accuracy of the stronger Qwen3-32B base model while achieving the highest recall overall. 
Although \method{} uses more searches than other 7B agents, this behavior is appropriate for a harder deep research setting: instead of enforcing a uniformly frugal policy, \method{} preserves retrieval when it believes more evidence is needed, yielding stronger zero-shot accuracy and recall.

\begin{table}[t]
\centering
\resizebox{0.8\columnwidth}{!}{
\begin{tabular}{llcc}
\toprule
Setting & Method & Acc. & Searches \\
\midrule
\multicolumn{4}{l}{\textit{\textbf{RL Optimizer on Qwen2.5-3B-Instruct}}} \\
GRPO & GiGPO & 39.8 & 1.32 \\
GRPO & \method{} & {41.7} & 1.60 \\
DAPO & GiGPO & 44.8 & 2.31 \\
DAPO & \method{} & {45.7} & 2.51 \\
PPO  & \method{} & 41.1 & 1.45 \\
\midrule
\multicolumn{4}{l}{\textit{\textbf{Backbone and Model Size with GRPO}}} \\
Llama3.2-3B & GiGPO & 41.7 & 1.05 \\
Llama3.2-3B & \method{} & {42.9} & 1.18 \\
Qwen3-14B & GiGPO & 45.6 & 1.05 \\
Qwen3-14B & \method{} & {46.5} & 1.36 \\
\bottomrule
\end{tabular}}
\caption{\textbf{Robustness across training settings.} Results are averaged over seven QA benchmarks. Llama3.2-3B denotes Llama-3.2-3B-Instruct.}
\label{tab:robustness}
\end{table}

\subsection{Robustness across Training Settings}
Table~\ref{tab:robustness} examines whether the gains of \method{} depend on a specific optimizer or backbone. 
With the default GRPO optimizer, \method{} improves over GiGPO by 1.9 points on Qwen2.5-3B-Instruct. 
The advantage also holds under DAPO~\citep{yu2025dapo}, where \method{} further improves accuracy from 44.8\% to 45.7\%. 
Since GiGPO is itself a group-based RL method and does not have a direct PPO~\citep{schulman2017proximal} counterpart, we additionally instantiate \method{} with PPO as an unpaired robustness check; the resulting accuracy remains competitive at 41.1\%. 
Across model families and sizes, \method{} consistently outperforms GiGPO on both Llama3.2-3B and Qwen3-14B. 
These results indicate that belief-action alignment functions as a robust reward-design principle for agentic RAG, rather than an optimization artifact of the default GRPO/Qwen2.5 setting. Detailed per-dataset results are provided in Appendix~\ref{sec:appendix_drar}.

\vspace{-0.17em}

\section{Conclusion}

We presented \method{}, a belief-action aligned training framework for agentic RAG. Instead of relying on external process supervision, \method{} probes the policy model's own answerability belief and compares it with each Search/Answer action. The resulting correctness-gated consistency reward provides lightweight training-time guidance for both over-search and under-search, without inference-time belief probing. Experiments on seven QA benchmarks show that \method{} consistently improves the accuracy--efficiency trade-off over strong RL-based agentic RAG baselines. Further analyses indicate that the gains are associated with better calibrated search behavior. \method{} mitigates premature answering, strengthens knowledge-boundary awareness, and transfers to harder deep research settings. Additional robustness experiments across optimizers and model backbones suggest that belief-action alignment is a robust reward-design principle for training search agents, rather than an artifact of a specific optimization or model setup.

\section*{Limitations}
\method{} introduces additional training-time computation. Although the belief probe is lightweight and avoids external LLM judges, it still requires an extra forward pass during reward calculation, and verify-first reasoning also increases generation length. As detailed in Appendix~\ref{sec:appendix_efficiency}, \method{} is substantially faster than HiPRAG in training, but remains slower than GiGPO. At inference time, the belief probe is not used, but verify-first reasoning can still produce longer per-turn responses and slightly higher latency than GiGPO.

Our experiments focus on QA-oriented agentic RAG with Search/Answer decisions. \method{} does not directly supervise the lexical quality of search queries, the faithfulness of generated rationales, or more complex tool-use actions beyond retrieval and answering. Extending belief-action alignment to richer agent settings with multiple tools, open-ended browsing actions, and long-horizon planning remains an important direction for future work.

\section*{Ethical Considerations}
In this work, we introduce a framework to improve the \search{}/\answer{} decision boundary of agentic RAG systems. While better retrieval decision-making can benefit knowledge-intensive applications, more capable search agents may also be misused for large-scale information gathering, surveillance, or misinformation generation. Our experiments are conducted on established public QA and research benchmarks, and do not involve personally identifiable information or private user data. Nevertheless, practical deployments should ensure that retrieval corpora are legally obtained, access-controlled when necessary, and paired with appropriate safeguards for high-stakes use cases.

\section*{Acknowledgements}
This work was supported by the National Natural Science Foundation of China under Grant 42394060 and 42394064, and Ant Group Research Fund.

\bibliography{emnlp2026_conference}
\clearpage
\appendix

\begin{table*}[t]
\centering

\resizebox{\textwidth}{!}{
\begin{tabular}{l*{3}{cc}|*{4}{cc}|cc}
\toprule
\multirow{3}{*}{\textbf{Belief Probe}}
& \multicolumn{6}{c|}{\textbf{Single-Hop QA}}
& \multicolumn{8}{c|}{\textbf{Multi-Hop QA}}
& \multicolumn{2}{c}{\multirow{2}{*}{\textbf{Avg.}}} \\
\cmidrule(lr){2-7}\cmidrule(lr){8-15}
& \multicolumn{2}{c}{NQ$^{\dagger}$}
& \multicolumn{2}{c}{TriviaQA$^{\star}$}
& \multicolumn{2}{c|}{PopQA$^{\star}$}
& \multicolumn{2}{c}{HotpotQA$^{\dagger}$}
& \multicolumn{2}{c}{2Wiki$^{\star}$}
& \multicolumn{2}{c}{MuSiQue$^{\star}$}
& \multicolumn{2}{c|}{Bamboogle$^{\star}$}
& \multicolumn{2}{c}{} \\
\cmidrule(lr){2-3}\cmidrule(lr){4-5}\cmidrule(lr){6-7}
\cmidrule(lr){8-9}\cmidrule(lr){10-11}\cmidrule(lr){12-13}
\cmidrule(lr){14-15}\cmidrule(lr){16-17}
& Acc. & Searches
& Acc. & Searches
& Acc. & Searches
& Acc. & Searches
& Acc. & Searches
& Acc. & Searches
& Acc. & Searches
& Acc. & Searches \\
\midrule

\multicolumn{17}{l}{\textit{\textbf{Qwen2.5-3B-Instruct}}} \\
$P(\mathrm{Yes})-P(\mathrm{No})$
& 45.6 & 1.17
& \best{62.8} & 1.19
& 48.0 & 1.27
& 41.3 & 1.58
& 39.1 & 1.87
& 15.9 & 2.24
& \best{39.5} & 1.90
& 41.7 & 1.60 \\

Internal Confidence
& \best{48.0} & 1.47
& 62.5 & 1.54
& \best{48.7} & 1.58
& \best{42.3} & 1.79
& \best{41.2} & 1.83
& \best{16.6} & 2.26
& 37.9 & 1.94
& \best{42.5} & 1.77 \\

\bottomrule
\end{tabular}
}
\caption{\textbf{Robustness to alternative belief probes on Qwen2.5-3B-Instruct.} We compare the default $P(\mathrm{Yes})-P(\mathrm{No})$ belief score with Internal Confidence, a stronger uncertainty estimator that aggregates yes/no self-evaluation signals across internal layers and token positions. For Internal Confidence, we use answerability threshold $\eta=0.5$. Acc. denotes Exact Match (EM) accuracy (\%), and Searches denotes the average number of search calls per question. Best accuracy values are \textbf{bold}.}
\label{tab:alt_probe}
\end{table*}

\begin{table*}[t]
\centering

\resizebox{\textwidth}{!}{
\begin{tabular}{l*{3}{cc}|*{4}{cc}|cc}
\toprule
\multirow{3}{*}{\textbf{Method}}
& \multicolumn{6}{c|}{\textbf{Single-Hop QA}}
& \multicolumn{8}{c|}{\textbf{Multi-Hop QA}}
& \multicolumn{2}{c}{\multirow{2}{*}{\textbf{Avg.}}} \\
\cmidrule(lr){2-7}\cmidrule(lr){8-15}
& \multicolumn{2}{c}{NQ$^{\dagger}$}
& \multicolumn{2}{c}{TriviaQA$^{\star}$}
& \multicolumn{2}{c|}{PopQA$^{\star}$}
& \multicolumn{2}{c}{HotpotQA$^{\dagger}$}
& \multicolumn{2}{c}{2Wiki$^{\star}$}
& \multicolumn{2}{c}{MuSiQue$^{\star}$}
& \multicolumn{2}{c|}{Bamboogle$^{\star}$}
& \multicolumn{2}{c}{} \\
\cmidrule(lr){2-3}\cmidrule(lr){4-5}\cmidrule(lr){6-7}
\cmidrule(lr){8-9}\cmidrule(lr){10-11}\cmidrule(lr){12-13}
\cmidrule(lr){14-15}\cmidrule(lr){16-17}
& Acc. & Searches
& Acc. & Searches
& Acc. & Searches
& Acc. & Searches
& Acc. & Searches
& Acc. & Searches
& Acc. & Searches
& Acc. & Searches \\
\midrule
\multicolumn{17}{l}{\textit{\textbf{Qwen2.5-3B-Instruct}}} \\

$\beta$-GRPO
& 39.3 & 1.28
& 59.3 & 0.90
& 35.7 & 1.50
& 31.6 & 2.45
& \best{39.1} & 2.85
& 10.7 & 2.50
& 34.4 & 2.15
& 35.7 & 2.15 \\

\method{}
& \textbf{45.6} & 1.17
& \textbf{62.8} & 1.19
& \textbf{48.0} & 1.27
& \textbf{41.3} & 1.58
& \textbf{39.1} & 1.87
& \textbf{15.9} & 2.24
& \textbf{39.5} & 1.90
& \textbf{41.7} & 1.60 \\

\bottomrule
\end{tabular}
}
\caption{\textbf{Comparison with confidence-threshold search training on Qwen2.5-3B-Instruct.} $\beta$-GRPO uses search-query token confidence to reward high-certainty search behavior, while \method{} uses answerability belief-action alignment to supervise Search/Answer decisions. Acc. denotes Exact Match (EM) accuracy (\%), and Searches denotes the average number of search calls per question. Best accuracy values are \textbf{bold}.}
\label{tab:search_wisely}
\end{table*}

\begin{table*}[t]
\centering
\resizebox{\textwidth}{!}{
\begin{tabular}{l*{3}{cc}|*{4}{cc}|cc}
\toprule
\multirow{3}{*}{\textbf{Method}}
& \multicolumn{6}{c|}{\textbf{Single-Hop QA}}
& \multicolumn{8}{c|}{\textbf{Multi-Hop QA}}
& \multicolumn{2}{c}{\multirow{2}{*}{\textbf{Avg.}}} \\
\cmidrule(lr){2-7}\cmidrule(lr){8-15}
& \multicolumn{2}{c}{NQ$^{\dagger}$}
& \multicolumn{2}{c}{TriviaQA$^{\star}$}
& \multicolumn{2}{c|}{PopQA$^{\star}$}
& \multicolumn{2}{c}{HotpotQA$^{\dagger}$}
& \multicolumn{2}{c}{2Wiki$^{\star}$}
& \multicolumn{2}{c}{MuSiQue$^{\star}$}
& \multicolumn{2}{c|}{Bamboogle$^{\star}$}
& \multicolumn{2}{c}{} \\
\cmidrule(lr){2-3}\cmidrule(lr){4-5}\cmidrule(lr){6-7}
\cmidrule(lr){8-9}\cmidrule(lr){10-11}\cmidrule(lr){12-13}
\cmidrule(lr){14-15}\cmidrule(lr){16-17}
& Acc. & Searches
& Acc. & Searches
& Acc. & Searches
& Acc. & Searches
& Acc. & Searches
& Acc. & Searches
& Acc. & Searches
& Acc. & Searches \\
\midrule

\multicolumn{17}{l}{\textit{\textbf{Qwen2.5-3B-Instruct}}} \\
GiGPO
& 45.1 & 1.00
& 60.9 & 0.99
& 44.9 & 1.17
& 38.0 & 1.24
& 37.1 & 1.55
& 14.5 & 1.81
& 37.9 & 1.49
& 39.8 & 1.32 \\

\method{} (GRPO)
& 45.6 & 1.17
& 62.8 & 1.19
& 48.0 & 1.27
& 41.3 & 1.58
& 39.1 & 1.87
& 15.9 & 2.24
& 39.5 & 1.90
& 41.7 & 1.60 \\

\method{} (GiGPO)
& \textbf{47.3} & 1.30
& \textbf{63.0} & 1.31
& \textbf{48.8} & 1.40
& \textbf{43.2} & 1.73
& \textbf{42.9} & 2.08
& \textbf{16.4} & 2.10
& \textbf{40.1} & 1.74
& \textbf{43.1} & 1.67 \\

\midrule
\multicolumn{17}{l}{\textit{\textbf{Qwen2.5-7B-Instruct}}} \\
GiGPO
& 46.8 & 0.96
& 65.8 & 0.71
& 48.1 & 1.14
& 42.3 & 1.20
& 42.9 & 1.40
& 18.2 & 2.09
& 43.2 & 1.22
& 43.9 & 1.25 \\

\method{} (GRPO)
& 46.9 & 1.62
& 67.2 & 1.19
& 48.5 & 1.85
& 45.7 & 1.67
& 44.9 & 1.69
& 20.3 & 2.31
& \textbf{44.0} & 1.83
& 45.4 & 1.74 \\

\method{} (GiGPO)
& \textbf{49.8} & 1.38
& \textbf{67.3} & 1.46
& \textbf{49.2} & 1.49
& \textbf{48.5} & 2.04
& \textbf{49.6} & 2.37
& \textbf{20.5} & 2.52
& \textbf{44.0} & 2.14
& \textbf{47.0} & 1.92 \\

\bottomrule
\end{tabular}
}
\caption{\textbf{Compatibility with GiGPO step-level credit assignment.} Results use the same setting as Table~\ref{tab:main}. $\dagger$ and $\star$ indicate in-domain and out-of-domain datasets, respectively. Acc. denotes Exact Match (EM) accuracy (\%), and Searches denotes the average number of search calls per question. Best accuracy values are \textbf{bold}.}
\label{tab:metarag_gigpo}
\end{table*}

\begin{table}[t]
\centering

\resizebox{0.8\columnwidth}{!}{
\begin{tabular}{lcc}
\toprule
Method & External Under-Search Rate (\%) \\
\midrule
\multicolumn{2}{l}{\textit{\textbf{Qwen2.5-3B-Instruct}}} \\
GiGPO & 35.6 \\
\method{} & 30.1 \\
\bottomrule
\end{tabular}
}
\caption{\textbf{External LLM-as-judge validation of under-search rate on Qwen2.5-3B-Instruct 400-step checkpoints.} We evaluate the full set of test trajectories using GLM-5.1. Lower is better. The judge is used only for post-hoc assessment and is not involved in reward computation, training, or model selection.}
\label{tab:external_undersearch}
\end{table}

\begin{table*}[t]
\centering

\resizebox{\textwidth}{!}{
\begin{tabular}{l*{3}{cc}|*{4}{cc}|cc}
\toprule
\multirow{3}{*}{\textbf{Variant}}
& \multicolumn{6}{c|}{\textbf{Single-Hop QA}}
& \multicolumn{8}{c|}{\textbf{Multi-Hop QA}}
& \multicolumn{2}{c}{\multirow{2}{*}{\textbf{Avg.}}} \\
\cmidrule(lr){2-7}\cmidrule(lr){8-15}
& \multicolumn{2}{c}{NQ$^{\dagger}$}
& \multicolumn{2}{c}{TriviaQA$^{\star}$}
& \multicolumn{2}{c|}{PopQA$^{\star}$}
& \multicolumn{2}{c}{HotpotQA$^{\dagger}$}
& \multicolumn{2}{c}{2Wiki$^{\star}$}
& \multicolumn{2}{c}{MuSiQue$^{\star}$}
& \multicolumn{2}{c|}{Bamboogle$^{\star}$}
& \multicolumn{2}{c}{} \\
\cmidrule(lr){2-3}\cmidrule(lr){4-5}\cmidrule(lr){6-7}
\cmidrule(lr){8-9}\cmidrule(lr){10-11}\cmidrule(lr){12-13}
\cmidrule(lr){14-15}\cmidrule(lr){16-17}
& Acc. & Searches
& Acc. & Searches
& Acc. & Searches
& Acc. & Searches
& Acc. & Searches
& Acc. & Searches
& Acc. & Searches
& Acc. & Searches \\
\midrule

\multicolumn{17}{l}{\textit{\textbf{Core Components}}} \\
w/o Consistency Reward
& {46.5} & 2.03
& 62.4 & 2.04
& {48.1} & 2.08
& {41.7} & 2.12
& {41.0} & 2.20
& {16.0} & 2.34
& 37.9 & 2.14
& {41.9} & 2.14 \\

w/o Verify-before-Action
& 45.6 & 1.09
& 60.4 & 1.12
& 45.0 & 1.21
& 40.1 & 1.38
& 39.0 & 1.60
& 15.4 & 1.70
& 37.5 & 1.57
& 40.4 & 1.38 \\

\midrule
\multicolumn{17}{l}{\textit{\textbf{Diagnostic Signal}}} \\
Over-Search Only
& 46.3 & 1.05
& {62.8} & 1.05
& 46.4 & 1.17
& 40.1 & 1.31
& 40.3 & 1.59
& 13.9 & 1.74
& 39.5 & 1.41
& 41.3 & 1.33 \\

Under-Search Only
& {46.9} & 2.33
& 62.1 & 2.22
& 47.9 & 2.40
& 41.2 & 2.53
& {42.3} & 2.85
& {15.9} & 2.89
& {40.3} & 2.70
& {42.4} & 2.56 \\

w/ Incorrect Trajectories
& 45.5 & 1.00
& 61.4 & 1.00
& 46.3 & 1.00
& 36.2 & 1.10
& 37.8 & 1.25
& 11.4 & 1.19
& 38.3 & 1.32
& 39.6 & 1.12 \\

\midrule
\multicolumn{17}{l}{\textit{\textbf{Candidate Strategy}}} \\
Always \search{}
& 44.7 & 1.33
& 62.0 & 1.49
& 47.6 & 1.46
& 41.0 & 1.83
& 39.7 & 2.01
& 15.1 & 2.37
& 37.9 & 1.93
& 41.1 & 1.77 \\

Always \answer{}
& 45.6 & 1.15
& {62.6} & 1.20
& 46.4 & 1.29
& 40.6 & 1.60
& 38.4 & 2.04
& 15.3 & 2.24
& 39.1 & 1.83
& 41.1 & 1.62 \\

Heuristic
& 44.8 & 1.15
& 62.2 & 1.13
& 46.6 & 1.18
& 40.5 & 1.47
& 38.9 & 1.68
& 14.4 & 1.94
& {41.1} & 1.70
& 41.2 & 1.46 \\

\midrule
\textbf{w/ \method{}}
& 45.6 & 1.17
& {62.8} & 1.19
& {48.0} & 1.27
& {41.3} & 1.58
& 39.1 & 1.87
& {15.9} & 2.24
& 39.5 & 1.90
& 41.7 & 1.60 \\

\bottomrule
\end{tabular}
}
\caption{\textbf{Detailed ablation study results on Qwen2.5-3B-Instruct.} $\dagger$ and $\star$ indicate in-domain and out-of-domain datasets, respectively. Acc. denotes Exact Match (EM) accuracy (\%), and Searches denotes the average number of search calls per question.}
\label{tab:ablation_full}
\end{table*}

\begin{table*}[t]
\centering

\resizebox{\textwidth}{!}{
\begin{tabular}{l*{3}{cc}|*{4}{cc}|cc}
\toprule
\multirow{3}{*}{\textbf{Method}}
& \multicolumn{6}{c|}{\textbf{Single-Hop QA}}
& \multicolumn{8}{c|}{\textbf{Multi-Hop QA}}
& \multicolumn{2}{c}{\multirow{2}{*}{\textbf{Avg.}}} \\
\cmidrule(lr){2-7}\cmidrule(lr){8-15}
& \multicolumn{2}{c}{NQ$^{\dagger}$}
& \multicolumn{2}{c}{TriviaQA$^{\star}$}
& \multicolumn{2}{c|}{PopQA$^{\star}$}
& \multicolumn{2}{c}{HotpotQA$^{\dagger}$}
& \multicolumn{2}{c}{2Wiki$^{\star}$}
& \multicolumn{2}{c}{MuSiQue$^{\star}$}
& \multicolumn{2}{c|}{Bamboogle$^{\star}$}
& \multicolumn{2}{c}{} \\
\cmidrule(lr){2-3}\cmidrule(lr){4-5}\cmidrule(lr){6-7}
\cmidrule(lr){8-9}\cmidrule(lr){10-11}\cmidrule(lr){12-13}
\cmidrule(lr){14-15}\cmidrule(lr){16-17}
& Acc. & Searches
& Acc. & Searches
& Acc. & Searches
& Acc. & Searches
& Acc. & Searches
& Acc. & Searches
& Acc. & Searches
& Acc. & Searches \\
\midrule

\multicolumn{17}{l}{\textit{\textbf{Qwen2.5-3B-Instruct}}} \\
GiGPO
& {45.1} & 1.00
& 60.9 & 0.99
& 44.9 & 1.17
& 38.0 & 1.24
& 37.1 & 1.55
& 14.5 & 1.81
& 37.9 & 1.49
& 39.8 & 1.32 \\

GiGPO w/ VF-Inf
& 44.7 & 1.04
& 60.8 & 1.03
& 44.4 & 1.17
& 38.4 & 1.28
& 36.6 & 1.63
& 13.3 & 1.85
& 37.2 & 1.52
& 39.3 & 1.36 \\

\method{}
& {45.6} & 1.17
& {62.8} & 1.19
& {48.0} & 1.27
& {41.3} & 1.58
& {39.1} & 1.87
& {15.9} & 2.24
& {39.5} & 1.90
& {41.7} & 1.60 \\

\method{} w/o VF-Inf
& 44.9 & 1.09
& {62.5} & 1.13
& {47.8} & 1.16
& {40.8} & 1.46
& {40.4} & 1.79
& {15.2} & 1.88
& {39.1} & 1.65
& {41.5} & 1.45 \\

\bottomrule
\end{tabular}
}
\caption{\textbf{Effect of verify-first inference on Qwen2.5-3B-Instruct.} VF-Inf denotes using Verify-first Action Generation at inference time. Acc. denotes Exact Match (EM) accuracy (\%), and Searches denotes the average number of search calls per question.}
\label{tab:verify_inference}
\end{table*}

\begin{table*}[t]
\centering

\resizebox{\textwidth}{!}{
\begin{tabular}{l*{3}{cc}|*{4}{cc}|cc}
\toprule
\multirow{3}{*}{\textbf{Setting}}
& \multicolumn{6}{c|}{\textbf{Single-Hop QA}}
& \multicolumn{8}{c|}{\textbf{Multi-Hop QA}}
& \multicolumn{2}{c}{\multirow{2}{*}{\textbf{Avg.}}} \\
\cmidrule(lr){2-7}\cmidrule(lr){8-15}
& \multicolumn{2}{c}{NQ$^{\dagger}$}
& \multicolumn{2}{c}{TriviaQA$^{\star}$}
& \multicolumn{2}{c|}{PopQA$^{\star}$}
& \multicolumn{2}{c}{HotpotQA$^{\dagger}$}
& \multicolumn{2}{c}{2Wiki$^{\star}$}
& \multicolumn{2}{c}{MuSiQue$^{\star}$}
& \multicolumn{2}{c|}{Bamboogle$^{\star}$}
& \multicolumn{2}{c}{} \\
\cmidrule(lr){2-3}\cmidrule(lr){4-5}\cmidrule(lr){6-7}
\cmidrule(lr){8-9}\cmidrule(lr){10-11}\cmidrule(lr){12-13}
\cmidrule(lr){14-15}\cmidrule(lr){16-17}
& Acc. & Searches
& Acc. & Searches
& Acc. & Searches
& Acc. & Searches
& Acc. & Searches
& Acc. & Searches
& Acc. & Searches
& Acc. & Searches \\
\midrule

\multicolumn{17}{l}{\textit{\textbf{Consistency Reward Weight $\lambda$}}} \\
$\lambda=0.0$
& {46.5} & 2.03
& 62.4 & 2.04
& {48.1} & 2.08
& {41.7} & 2.12
& {41.0} & 2.20
& {16.0} & 2.34
& 37.9 & 2.14
& {41.9} & 2.14 \\

$\lambda=0.1$ (default)
& 45.6 & 1.17
& {62.8} & 1.19
& 48.0 & 1.27
& {41.3} & 1.58
& 39.1 & 1.87
& {15.9} & 2.24
& {39.5} & 1.90
& {41.7} & 1.60 \\

$\lambda=0.2$
& 44.5 & 1.22
& 61.9 & 1.20
& 45.3 & 1.38
& 40.4 & 1.60
& 38.9 & 2.08
& 15.7 & 2.15
& {40.3} & 1.80
& 41.0 & 1.63 \\

$\lambda=0.3$
& 44.2 & 1.03
& 61.4 & 1.05
& 45.6 & 1.06
& 38.2 & 1.26
& 37.2 & 1.52
& 13.3 & 1.48
& 38.3 & 1.37
& 39.8 & 1.25 \\

$\lambda=0.4$
& {46.7} & 1.16
& {62.7} & 1.29
& {48.4} & 1.20
& 40.5 & 1.61
& {39.7} & 1.67
& 15.5 & 2.08
& 37.5 & 1.74
& 41.6 & 1.54 \\

\midrule
\multicolumn{17}{l}{\textit{\textbf{Margin $m$}}} \\
$m=0.0$ (default)
& 45.6 & 1.17
& {62.8} & 1.19
& {48.0} & 1.27
& {41.3} & 1.58
& {39.1} & 1.87
& {15.9} & 2.24
& {39.5} & 1.90
& {41.7} & 1.60 \\

$m=0.2$
& 44.6 & 1.39
& 61.5 & 1.52
& 46.4 & 1.59
& 39.8 & 1.82
& 38.3 & 2.04
& {15.3} & 2.39
& {41.2} & 2.07
& 41.0 & 1.83 \\

$m=0.4$
& {45.7} & 1.03
& 61.8 & 1.05
& 46.3 & 1.05
& 38.5 & 1.25
& 36.8 & 1.42
& 12.3 & 1.57
& 38.3 & 1.65
& 39.9 & 1.29 \\

$m=0.6$
& 45.3 & 1.10
& 61.5 & 1.13
& {46.6} & 1.21
& 39.2 & 1.51
& 38.2 & 1.74
& 15.1 & 2.01
& 38.7 & 1.70
& 40.7 & 1.49 \\

$m=0.8$
& {46.5} & 1.04
& {62.1} & 1.08
& 46.2 & 1.10
& {40.7} & 1.42
& {42.1} & 1.61
& 14.2 & 1.89
& 37.5 & 1.59
& {41.3} & 1.39 \\

\bottomrule
\end{tabular}
}
\caption{\textbf{Detailed hyperparameter sensitivity results on Qwen2.5-3B-Instruct.} $\dagger$ and $\star$ indicate in-domain and out-of-domain datasets, respectively. Acc. denotes Exact Match (EM) accuracy (\%), and Searches denotes the average number of search calls per question.}
\label{tab:hyper_full}
\end{table*}

\begin{table*}[t]
\centering
\resizebox{0.8\textwidth}{!}{
\begin{tabular}{l*{4}{cc}}
\toprule
\multirow{2}{*}{\textbf{Method}}
& \multicolumn{2}{c}{\textbf{GSM8K}}
& \multicolumn{2}{c}{\textbf{SciQ}}
& \multicolumn{2}{c}{\textbf{TriviaQA}}
& \multicolumn{2}{c}{\textbf{Avg.}} \\
\cmidrule(lr){2-3}\cmidrule(lr){4-5}\cmidrule(lr){6-7}\cmidrule(lr){8-9}
& AUROC & PRR & AUROC & PRR & AUROC & PRR & AUROC & PRR \\
\midrule

\multicolumn{9}{l}{\textit{\textbf{Perplexity}}} \\
Base
& 56.0 & \best{16.1}
& \second{60.1} & \second{22.7}
& 62.5 & 22.6
& \second{59.5} & \second{20.5} \\

Search-R1
& 54.4 & 10.1
& 59.8 & 22.3
& 61.4 & 21.2
& 58.5 & 17.9 \\

HiPRAG
& 52.2 & 2.7
& 57.9 & 18.9
& 62.4 & 24.4
& 57.5 & 15.3 \\

GiGPO
& \second{56.2} & 13.8
& 56.8 & 14.6
& \second{63.7} & \second{26.7}
& 58.9 & 18.4 \\

\method{}
& \best{56.3} & \second{15.5}
& \best{60.2} & \best{22.9}
& \best{64.0} & \best{27.1}
& \best{60.2} & \best{21.8} \\

\midrule
\multicolumn{9}{l}{\textit{\textbf{Attentional Entropy}}} \\
Base
& \second{55.6} & \second{15.3}
& 56.3 & 15.2
& 57.9 & 13.1
& 56.6 & \second{14.5} \\

Search-R1
& 54.0 & 9.4
& \best{56.9} & \best{16.9}
& 58.1 & 14.6
& 56.3 & 13.6 \\

HiPRAG
& 54.3 & 6.9
& 54.9 & 13.2
& {58.9} & {17.3}
& 56.0 & 12.5 \\

GiGPO
& 55.0 & 11.5
& 55.0 & 11.1
& \second{60.0} & \second{19.3}
& \second{56.7} & 14.0 \\

\method{}
& \best{55.8} & \best{15.5}
& \best{56.9} & \second{16.1}
& \best{60.6} & \best{19.8}
& \best{57.8} & \best{17.1} \\

\bottomrule
\end{tabular}
}
\caption{\textbf{Detailed attribution analysis results on Qwen2.5-7B-Instruct.} We evaluate knowledge-boundary awareness using perplexity and attentional entropy as diagnostic signals, and report AUROC and PRR on GSM8K, SciQ, and TriviaQA. Avg. denotes the average over the three datasets. Best and second-best values are \textbf{bold} and \underline{underlined} within each diagnostic-signal block.}
\label{tab:attribution_full}
\end{table*}

\begin{table*}[t]
\centering
\resizebox{\textwidth}{!}{
\begin{tabular}{ll*{3}{cc}|*{4}{cc}|cc}
\toprule
\multirow{3}{*}{\textbf{Setting}}
& \multirow{3}{*}{\textbf{Method}}
& \multicolumn{6}{c|}{\textbf{Single-Hop QA}}
& \multicolumn{8}{c|}{\textbf{Multi-Hop QA}}
& \multicolumn{2}{c}{\multirow{2}{*}{\textbf{Avg.}}} \\
\cmidrule(lr){3-8}\cmidrule(lr){9-16}
& &
\multicolumn{2}{c}{NQ$^{\dagger}$}
& \multicolumn{2}{c}{TriviaQA$^{\star}$}
& \multicolumn{2}{c|}{PopQA$^{\star}$}
& \multicolumn{2}{c}{HotpotQA$^{\dagger}$}
& \multicolumn{2}{c}{2Wiki$^{\star}$}
& \multicolumn{2}{c}{MuSiQue$^{\star}$}
& \multicolumn{2}{c|}{Bamboogle$^{\star}$}
& \multicolumn{2}{c}{} \\
\cmidrule(lr){3-4}\cmidrule(lr){5-6}\cmidrule(lr){7-8}
\cmidrule(lr){9-10}\cmidrule(lr){11-12}\cmidrule(lr){13-14}
\cmidrule(lr){15-16}\cmidrule(lr){17-18}
& & Acc. & Searches
& Acc. & Searches
& Acc. & Searches
& Acc. & Searches
& Acc. & Searches
& Acc. & Searches
& Acc. & Searches
& Acc. & Searches \\
\midrule

\multicolumn{18}{l}{\textit{\textbf{RL Optimizer on Qwen2.5-3B-Instruct}}} \\
GRPO & GiGPO
& 45.1 & 1.00
& 60.9 & 0.99
& 44.9 & 1.17
& 38.0 & 1.24
& 37.1 & 1.55
& 14.4 & 1.81
& 37.9 & 1.49
& 39.8 & 1.32 \\

GRPO & \method{}
& 45.6 & 1.17
& 62.8 & 1.19
& 48.0 & 1.27
& 41.3 & 1.58
& 39.1 & 1.87
& 15.9 & 2.24
& 39.5 & 1.90
& 41.7 & 1.60 \\

DAPO & GiGPO
& 47.3 & 2.01
& 63.6 & 1.91
& 48.3 & 2.13
& 46.2 & 2.30
& 46.6 & 2.56
& 19.8 & 2.78
& 42.0 & 2.46
& 44.8 & 2.31 \\

DAPO & \method{}
& 48.6 & 2.33
& 63.9 & 2.26
& 49.9 & 2.37
& 46.1 & 2.42
& 48.3 & 2.70
& 18.9 & 2.84
& 44.0 & 2.62
& 45.7 & 2.51 \\

PPO & \method{}
& 45.7 & 1.08
& 62.2 & 1.13
& 46.4 & 1.15
& 40.4 & 1.44
& 40.1 & 1.75
& 14.4 & 1.88
& 38.3 & 1.72
& 41.1 & 1.45 \\

\midrule
\multicolumn{18}{l}{\textit{\textbf{Backbone and Model Size with GRPO}}} \\
Llama3.2-3B & GiGPO
& 45.3 & 0.67
& 62.7 & 0.54
& 46.3 & 0.93
& 39.3 & 1.00
& 39.7 & 1.30
& 14.5 & 1.93
& 44.0 & 0.99
& 41.7 & 1.05 \\

Llama3.2-3B & \method{}
& 46.2 & 0.77
& 64.4 & 0.69
& 45.7 & 0.91
& 40.9 & 1.15
& 42.4 & 1.41
& 16.3 & 2.15
& 44.4 & 1.16
& 42.9 & 1.18 \\

Qwen3-14B & GiGPO
& 45.9 & 0.72
& 69.0 & 0.50
& 51.0 & 0.88
& 42.3 & 1.07
& 46.2 & 1.52
& 19.7 & 1.55
& 44.8 & 1.08
& 45.6 & 1.05 \\

Qwen3-14B & \method{}
& 46.4 & 1.11
& 69.8 & 1.12
& 48.9 & 1.13
& 45.7 & 1.36
& 47.3 & 1.79
& 21.9 & 1.64
& 45.6 & 1.38
& 46.5 & 1.36 \\

\bottomrule
\end{tabular}
}
\caption{\textbf{Detailed robustness analysis results across training settings.} $\dagger$ and $\star$ indicate in-domain and out-of-domain datasets, respectively. Acc. denotes Exact Match (EM) accuracy (\%), and Searches denotes the average number of search calls per question. Llama3.2-3B denotes Llama-3.2-3B-Instruct.}
\label{tab:robustness_full}
\end{table*}

\begin{table}[t]
\centering

\resizebox{0.8\columnwidth}{!}{
\begin{tabular}{lccc}
\toprule
Method & Train / Step & Resp. Len. & Infer. / Query \\
\midrule
\multicolumn{4}{l}{\textit{\textbf{Qwen2.5-7B-Instruct}}} \\
HiPRAG & 525s & 151.4 & 0.1899s \\
GiGPO  & 249s & 104.6 & 0.1179s \\
\method{} & 281s & 119.7 & 0.1453s \\
\bottomrule
\end{tabular}
}
\caption{\textbf{Training and inference efficiency on Qwen2.5-7B-Instruct.} All measurements are averaged over runs on a single node with 8 A100 GPUs. Train / Step denotes the wall-clock time for one training step. Resp. Len. denotes the average response length per turn during agentic RAG inference. Infer. / Query denotes the average end-to-end inference time per query.}
\label{tab:efficiency}
\end{table}

\section{Robustness to Alternative Belief Probes}
\label{sec:appendix_alt_probe}

To examine whether \method{} is tied to the simple $P(\mathrm{Yes})-P(\mathrm{No})$ belief probe, we replace it with a stronger uncertainty estimator, Internal Confidence~\citep{chen2025query}. Unlike the last-token yes/no probability used in the main experiments, Internal Confidence aggregates self-evaluation signals across internal layers and token positions. Given its confidence score $u_t\in[0,1]$, we use a threshold $\eta=0.5$ to determine answerability: a step is treated as answerable when $u_t\ge\eta$ and insufficient otherwise. The consistency reward is then computed with the same belief-action alignment rule as in the main method.

Table~\ref{tab:alt_probe} shows that replacing the default belief probe with Internal Confidence improves the average accuracy from 41.7\% to 42.5\%. The gain is observed on most benchmarks, including NQ, PopQA, HotpotQA, 2Wiki, and MuSiQue. This indicates that the effectiveness of belief-action alignment is not specific to the last-token $P(\mathrm{Yes})-P(\mathrm{No})$ implementation. Stronger belief probes can be plugged into the same reward framework and may provide better supervision, at the cost of slightly more retrieval.

\section{Case Study: Belief-Action Gap Diagnosis}
\label{sec:appendix_case_bag_diagnosis}

We further examine whether the proposed Belief-Action Gap Diagnosis can identify both types of suboptimal retrieval behavior using cases from a GiGPO-trained Qwen2.5-3B-Instruct agent. With the default margin $m=0$, a positive Belief Score indicates that the model believes the current context is answerable, while a negative Belief Score indicates that more evidence is needed. Therefore, a positive belief followed by \search{} is diagnosed as \textsc{Over-search}, and a negative belief followed by \answer{} is diagnosed as \textsc{Under-search}.

\paragraph{Case: Over-Search.}
Figure~\ref{case:over_search_gigpo} shows an over-search trajectory. After the first retrieval, the correct answer is already present: Doc 2 identifies Toby Kebbell as an English actor known for \textit{Dead Man's Shoes} (2004) and Doc 3 confirms that he starred in \textit{Kong: Skull Island}. Despite a strongly positive Belief Score, the agent continues searching and pivots to Tom Hiddleston, eventually producing an unsupported wrong answer.

\paragraph{Case: Under-Search.}
Figure~\ref{case:under_search_gigpo_2} shows the opposite failure mode. The agent retrieves the guitarist's identity, Vivian Campbell, but not his nationality. The Belief Score is negative, indicating insufficient answerability, yet the agent answers by transferring the band's nationality to the guitarist. This is correctly diagnosed as \textsc{Under-search}.

\section{Pseudocode for Reward Calculation}
\label{sec:appendix_pseudocode}
Algorithm~\ref{alg:metarag} summarizes the reward calculation procedure of \method{}, covering Verify-first Action Generation, Internal Belief Probing, Belief-Action Gap Diagnosis, and Consistency Reward Calculation.

\begin{algorithm}[t]
\caption{Reward calculation in \method{}}
\label{alg:metarag}
\begin{algorithmic}[1]
\Require Question $q$, gold answer $y^*$, policy model $\pi_\theta$, retriever $\mathcal{R}$, margin $m$, weight $\lambda$
\State Rollout a trajectory $\tau=(c_1,a_1,\ldots,c_{T_\tau},a_{T_\tau})$ with Verify-first Action Generation.
\State Extract the final answer $\hat{y}$ and action types $\{\alpha_t\}_{t=1}^{T_\tau}$ from $\tau$.
\For{each decision step $t=1,\ldots,T_\tau$}
    \State Run Internal Belief Probing with Question \& History.
    \State Compute Belief Score $b_t=P_t(\mathrm{Yes})-P_t(\mathrm{No})$.
    \If{$b_t>m$ and $\alpha_t=\answer$}
        \State $o_t\leftarrow 1$ \Comment{\textsc{Optimal}}
    \ElsIf{$b_t<-m$ and $\alpha_t=\search$}
        \State $o_t\leftarrow 1$ \Comment{\textsc{Optimal}}
    \ElsIf{$b_t>m$ and $\alpha_t=\search$}
        \State $o_t\leftarrow 0$ \Comment{\textsc{Over-search}}
    \ElsIf{$b_t<-m$ and $\alpha_t=\answer$}
        \State $o_t\leftarrow 0$ \Comment{\textsc{Under-search}}
    \Else
        \State $o_t\leftarrow 0$ \Comment{Margin Zone}
    \EndIf
\EndFor
\State $R_{{consist}}\leftarrow T_\tau^{-1}\sum_{t=1}^{T_\tau}o_t$.
\State $R_{{outcome}}\leftarrow \mathbb{I}[\mathrm{EM}(\hat{y},y^*)]$.
\State $R_{{total}}\leftarrow R_{{outcome}} \cdot (1+\lambda R_{{consist}})$.
\State \Return $R_{{total}}$ for downstream RL optimization.
\end{algorithmic}
\end{algorithm}

\section{Experiment Details}
\label{appendix:ed}
\subsection{Details of Training}
\label{appendix:td}
\paragraph{Hyperparameters.}
The maximum prompt length is 4096 tokens, and the maximum response length is 512 tokens. The max turn is set to 4. The learning rate is 1e-6 for the actor. We adopt a rule-based reward, assigning a reward of 1 for success and 0 for failure. Invalid actions are penalized with a reward of -0.01. We set the training batch size to 256 and use a rollout group size of 5. Rollout and validation temperatures are set to 1.0 and 0.0, respectively. The mini-batch size is 512, and the KL penalty coefficient is set to $0.001$. Unless otherwise specified, the candidate action type $\tilde{\alpha}_t$ is sampled uniformly from $\{\search,\answer\}$, the consistency reward weight $\lambda$ is set to $0.1$, and the margin $m$ is set to $0.0$.

\paragraph{Computing Details.}
Qwen2.5-3B-Instruct uses 4×A100 GPUs and Qwen2.5-7B-Instruct uses 8×A100 GPUs, each for 400 iterations.

\subsection{Prompts}
The prompts we use for \method{} are presented in Figures~\ref{prompt:search_verify} and~\ref{prompt:search_probe}. These prompt templates are constructed using Python-style string formatting, where placeholders enclosed in curly braces (\{\}) represent semantic slots. These placeholders, such as \{task\_description\}, \{step\_count\}, and \{memory\_context\}, are dynamically populated at runtime via Python's \texttt{.format()} function. 

The search agent outputs reasoning traces within <think> </think>, issues search queries within <search> </search>, and provides anwsers within <answer> </answer>. Retrieved evidence from the retriever is presented in <information> </information> tags.

\section{Comparison with Confidence-Threshold Search Training}
\label{sec:appendix_search_wisely}
Table~\ref{tab:search_wisely} compares \method{} with $\beta$-GRPO~\citep{wu2025search}, a confidence-threshold training method that rewards correct trajectories only when the generated search queries exceed a confidence threshold. 
Under the same Qwen2.5-3B-Instruct setting, \method{} improves the average accuracy from 35.7\% to 41.7\% while reducing the average number of searches from 2.15 to 1.60. 
The gain is especially clear on multi-hop benchmarks, where deciding whether the current evidence is sufficient is more important than merely encouraging high-confidence search queries. 
These results suggest that directly aligning answerability belief with the Search/Answer action provides a stronger decision-level signal than thresholding search-query confidence alone.

\section{Compatibility with Step-level Credit Assignment}
\label{sec:appendix_gigpo}

We further combine \method{} with GiGPO's step-level credit assignment under the same setting as Table~\ref{tab:main}. As shown in Table~\ref{tab:metarag_gigpo}, \method{} (GiGPO) improves the average accuracy over GiGPO from 39.8\% to 43.1\% on Qwen2.5-3B-Instruct and from 43.9\% to 47.0\% on Qwen2.5-7B-Instruct, corresponding to gains of 3.3 and 3.1 points, respectively. This indicates that belief-action alignment is complementary to stronger step-level credit assignment.

\section{External Validation of Under-Search Rate}
\label{sec:appendix_external_undersearch}

The under-search rate in Figure~\ref{fig:dynamics} is computed from the internal belief-based diagnosis used during training. This diagnostic is useful for monitoring whether the policy increasingly aligns its \search{}/\answer{} decisions with its own answerability belief, but it may raise a circularity concern because the same belief signal also contributes to the consistency reward. We therefore conduct an external validation using an LLM-as-judge protocol on the full set of test trajectories generated by the 400-step checkpoints of GiGPO and \method{}.

Our protocol is inspired by judge-based process diagnosis in HiPRAG~\citep{wu2025hiprag}, but differs in two important ways. First, the external judge is used only for post-hoc evaluation, not for reward computation, training, or model selection. Second, instead of judging isolated non-search steps with only step-local reasoning, we evaluate the terminal \answer{} decision in the full agentic RAG context: the original question, retrieved history, model reasoning, final answer, and gold answer. This directly targets premature answering in our \search{}/\answer{} boundary setting.

\paragraph{Evaluation Protocol.}
For each evaluated trajectory, we extract the final \answer{} step and provide the judge with: (1) the question, (2) all retrieved passages available before the answer, (3) the model's reasoning immediately before answering, (4) the final answer, and (5) the gold answer. The judge determines whether the agent answered prematurely because the evidence was insufficient, missing a necessary entity or attribute, or misused in a way that should have triggered another search before answering. We mark a trajectory as under-search only when the judge concludes that the answer step should have been preceded by an additional search. Cases where the evidence is sufficient but the model reasons incorrectly are not counted as under-search.

We use GLM-5.1 as the external judge with deterministic decoding (temperature $=0$), using the prompt template shown in Figure~\ref{prompt:under_external_judge}. The judge is not given the internal Belief Score, the candidate action, or any reward information.

As shown in Table~\ref{tab:external_undersearch}, \method{} reduces the externally judged under-search rate from 35.6\% to 30.1\%, a 5.5-point absolute reduction over GiGPO. This supports the trend observed in Figure~\ref{fig:dynamics} and indicates that the reduction in premature answering is not merely an artifact of the internal belief probe. The absolute rates are lower than the belief-based diagnostic because the external judge uses a stricter criterion: it only marks cases where the terminal answer should clearly have been preceded by an additional search. Unlike judge-based process-supervised methods such as HiPRAG, \method{} does not require external LLM judges during training, preserving the scalability advantage of belief-action alignment.

\section{Case Study of Under-Search Mitigation}
\label{sec:appendix_case_under_search_mitigation}

To illustrate how \method{} mitigates premature answering, we compare GiGPO and \method{}, both based on Qwen2.5-3B-Instruct, on the same HotpotQA question. The question requires the agent to first identify the Earl associated with Mold Castle and then resolve the name by which the Earl was also known. As shown in Figures~\ref{case:under_search_gigpo} and~\ref{case:under_search_metarag}, GiGPO stops after the first retrieval and mistakes the title ``Earl of Chester'' for the requested alias, while \method{} rejects a premature \answer{} candidate and issues a targeted follow-up search.

\paragraph{Case: GiGPO Under-Search.}
Figure~\ref{case:under_search_gigpo} shows that GiGPO retrieves the relevant entity, Hugh d'Avranches, Earl of Chester, but does not further verify the alias. The agent therefore answers with the title rather than the alternative name, resulting in an under-search error.

\paragraph{Case: \method{} Avoids Under-Search.}
Figure~\ref{case:under_search_metarag} shows the corresponding \method{} trajectory. At Step 2, the proposed candidate action is \answer{}, but the agent identifies that the current evidence only resolves the Earl's identity and title. \method{} therefore overrides the candidate and searches for the missing alias, ultimately retrieving and answering ``Hugh the Fat''.

\section{Detailed Ablation Study Results}
\label{sec:appendix_dasr}
Table~\ref{tab:ablation_full} provides the per-dataset breakdown for the ablation study results (corresponding to Table~\ref{tab:ablation} in the main paper).

\section{Effect of Verify-first Inference}
\label{sec:appendix_verify_inference}

Verify-first Action Generation changes the action-generation interface, so we further examine whether the gains come merely from using a stronger inference prompt. We consider two inference-time controls on Qwen2.5-3B-Instruct. First, we apply verify-first inference to a GiGPO-trained model, without changing its training. This tests whether the prompt format alone benefits a baseline policy. Second, we evaluate \method{} without verify-first inference, using the shorter standard action prompt at test time. This setting can be viewed as a training-only use of verify-first reasoning, without additional distillation.

Table~\ref{tab:verify_inference} shows that simply adding verify-first inference to GiGPO does not improve the baseline: its average accuracy decreases from 39.8\% to 39.3\%, while the number of searches slightly increases from 1.32 to 1.36. Thus, the improvement of \method{} cannot be explained by an inference-time prompt advantage alone.

For \method{}, disabling verify-first inference reduces the average accuracy only slightly, from 41.7\% to 41.5\%, while lowering the average number of searches from 1.60 to 1.45. This suggests that verify-first reasoning is useful during training and still provides a small benefit at inference, but \method{} largely retains its advantage even with a shorter standard inference prompt. Together with the prompt-matched ablation without consistency reward in Table~\ref{tab:ablation}, these results indicate that verify-first prompting and belief-action reward shaping play different roles: verify-first reasoning provides a better decision interface, while the consistency reward calibrates the \search{}/\answer{} boundary and improves the accuracy--efficiency trade-off.

\section{Detailed Hyperparameter Sensitivity Results}
\label{sec:appendix_dhsr}
Table~\ref{tab:hyper_full} provides the per-dataset breakdown for the hyperparameter sensitivity results (corresponding to Figure~\ref{fig:hyper} in the main paper).

\section{Detailed Attribution Analysis Results}
\label{sec:appendix_daar}

Table~\ref{tab:attribution_full} provides the per-dataset breakdown for the knowledge-boundary awareness attribution results (corresponding to Figure~\ref{fig:kb} in the main paper). In addition to the perplexity-based results reported in the main text, we also include attentional entropy~\citep{duan2024shifting} as an alternative diagnostic signal.

\section{Detailed Robustness Analysis Results}
\label{sec:appendix_drar}

Table~\ref{tab:robustness_full} provides the per-dataset breakdown for the robustness results across optimizers, model families, and model sizes (corresponding to Table~\ref{tab:robustness} in the main paper).

\section{Training and Inference Efficiency}
\label{sec:appendix_efficiency}

Table~\ref{tab:efficiency} reports the training and inference efficiency of representative RL-based agentic RAG methods. Compared with HiPRAG, which relies on external LLM-based process diagnosis, \method{} reduces the training time per step from 525s to 281s. This shows that using the policy model's own belief signal provides substantially lighter process supervision than judge-based diagnosis.

Compared with GiGPO, \method{} introduces moderate additional training cost, increasing the time per step from 249s to 281s. This overhead comes from verify-first reasoning and the training-time belief probe. At inference time, however, the belief probe is not used. As a result, \method{} remains efficient, requiring 0.1453s per query on average, between GiGPO and HiPRAG. These results indicate that \method{} improves decision supervision with manageable computational overhead, while avoiding the expensive external-diagnosis pipeline used by judge-based process supervision.

\clearpage
\onecolumn

\begin{figure}[t]
\centering
\resizebox{\textwidth}{!}{
\begin{tcolorbox}[colback=gray!5!white, colframe=black!75!black,
title=Case: Over-Search Diagnosed by Belief-Action Gap Diagnosis, boxrule=0.3mm, width=\textwidth, arc=3mm, auto outer arc=true]

\textbf{Question:} Kong: Skull Island starred the English actor known for his roles in what 2004 film? \\
\textbf{Ground Truth:} Dead Man's Shoes \\[6pt]

\textbf{Step 1}

\textbf{Reasoning:} I need to find out which 2004 film the English actor starred in that was also in the 2012 film ``Kong: Skull Island''. I will search the web to find this information.

\textbf{Action:} \texttt{search(``English actor starred in 2004 film also in Kong: Skull Island'')} \\

\textbf{Step 2}

\textbf{Observation:} [Doc 1] Andy Serkis \ldots\ is an English actor \ldots\ King Kong in the eponymous 2005 film \ldots\ [Doc 2] \textbf{Toby Kebbell} \ldots\ is an English stage and film actor. He is known for his roles in films such as \textbf{``Dead Man's Shoes'' (2004)} \ldots\ He most recently starred in the second film of the MonsterVerse film series [i.e., Kong: Skull Island]. [Doc 3] Kong: Skull Island \ldots\ The film stars Tom Hiddleston, \ldots, \textbf{Toby Kebbell}, \ldots

\textbf{Belief Probing:} $P(\mathrm{Yes}) = 0.9990$, $P(\mathrm{No}) = 0.0010$, Belief Score $= +0.9980$

\textbf{Reasoning:} From the information provided, we can see that Kong: Skull Island starred Tom Hiddleston. To find out which 2004 film Tom Hiddleston was known for, we need to look for his roles in 2004. \ldots\ To confirm, we should search for more specific information about Tom Hiddleston's roles in 2004.

\textbf{Action:} \underline{\texttt{search(``Tom Hiddleston roles 2004'')}} \quad \textit{$\leftarrow$ Diagnosed as \textsc{Over-search}: despite a strongly positive belief score ($b_t=+0.9980$), the agent chooses \search{}, ignoring the sufficient evidence in Doc 2 that links Toby Kebbell to ``Dead Man's Shoes'' and incorrectly pivoting to Tom Hiddleston.} \\

\textbf{Step 3}

\textbf{Observation:} [Doc 1--3] Tom Hiddleston \ldots\ appeared in West End theatre productions of ``Cymbeline'' (2007) and ``Ivanov'' (2008) \ldots\ came to wider public attention when cast as Loki \ldots\ ``Thor'' (2011), ``The Avengers'' (2012) \ldots\ TV credits include ``Wallander'' (2008), ``Suburban Shootout'' (2006), ``Miss Austen Regrets'' (2008) \ldots

\textbf{Belief Probing:} $P(\mathrm{Yes}) = 0.9994$, $P(\mathrm{No}) = 0.0006$, Belief Score $= +0.9988$

\textbf{Reasoning:} From the information provided, we can see that Tom Hiddleston starred in the film ``Thor'' in 2011, not 2004. \ldots\ The information provided does not indicate any 2004 film in which Tom Hiddleston starred. Therefore, we may need to search further to find the correct answer.

\textbf{Action:} \underline{\texttt{search(``Tom Hiddleston films 2004'')}} \quad \textit{$\leftarrow$ Diagnosed as \textsc{Over-search}: despite a strongly positive belief score ($b_t=+0.9988$), the agent chooses \search{} again, issuing another redundant query while still pursuing the wrong actor.} \\

\textbf{Step 4}

\textbf{Observation:} [Doc 2] \ldots\ the same year he starred as Freddie Page, a RAF pilot in the drama \textbf{``The Deep Blue Sea''}, alongside Rachel Weisz. In 2012, he reprised his role as the supervillain Loki in ``The Avengers'' \ldots\ [Doc 3] In 2017, Hiddleston starred in Legendary Pictures' King Kong film \ldots\ directed by Jordan Vogt-Roberts.

\textbf{Reasoning:} From the information provided, we can see that Tom Hiddleston starred in the 2004 film ``The Deep Blue Sea''. \ldots\ Therefore, the 2004 film in question is ``The Deep Blue Sea''.

\textbf{Action:} \underline{\texttt{answer(``The Deep Blue Sea'')}} \quad \textit{$\leftarrow$ Wrong answer: the correct answer ``Dead Man's Shoes'' was already present after the first search}

\end{tcolorbox}
}
\caption{\textbf{Over-search diagnosed by Belief-Action Gap Diagnosis.} The correct answer \textit{Dead Man's Shoes} is already available after the first retrieval, and the Belief Score is strongly positive, but the agent continues searching and eventually returns an incorrect answer.}
\label{case:over_search_gigpo}
\end{figure}

\begin{figure}[t]
\centering
\resizebox{\textwidth}{!}{
\begin{tcolorbox}[colback=gray!5!white, colframe=black!75!black,
title=Case: Under-Search Diagnosed by Belief-Action Gap Diagnosis, boxrule=0.3mm, width=\textwidth, arc=3mm, auto outer arc=true]

\textbf{Question:} Last in Line reunited in 2010 with the guitarist who was of what nationality? \\
\textbf{Ground Truth:} Northern Irish \\[6pt]

\textbf{Step 1}

\textbf{Reasoning:} I need to find out who the guitarist was and what nationality they were. I will search the web to find this information.

\textbf{Action:} \texttt{search(``who was the guitarist of Last in Line who reunited in 2010'')} \\

\textbf{Step 2}

\textbf{Observation:} [Doc 2] Last in Line is an \textbf{American} heavy metal band formed in 2012 by former members of the original lineup of Dio \ldots\ Following the death of Dio frontman Ronnie James Dio in 2010, the original lineup of drummer Vinny Appice, bassist Jimmy Bain, guitarist \textbf{Vivian Campbell}, and keyboardist Claude Schnell reunited \ldots\ [Doc 3] After the death of Ronnie James Dio in 2010, \textbf{Vivian Campbell} set into motion the idea of reuniting the original members of Dio for the first time since his departure from the band in 1986. \ldots

\textbf{Belief Probing:} $P(\mathrm{Yes}) = 0.3208$, $P(\mathrm{No}) = 0.6792$, Belief Score $= -0.3584$

\textbf{Reasoning:} From the information provided, we can see that Last in Line is an American heavy metal band. The question asks about the guitarist who reunited with Last in Line in 2010. The information states that after the death of Dio frontman Ronnie James Dio in 2010, Vivian Campbell set into motion the idea of reuniting the original members of Dio for the first time since his departure from the band in 1986. This implies that Vivian Campbell, who was the guitarist of the original Dio lineup, reunited with Last in Line in 2010. To confirm this, I will not need to search further as the information is directly provided in the information section.

\textbf{Action:} \underline{\texttt{answer(``American'')}} \quad \textit{$\leftarrow$ Diagnosed as \textsc{Under-search}: despite a negative belief score ($b_t=-0.3584$) signaling insufficient confidence, the agent chooses \answer{} without a follow-up search for Vivian Campbell's personal background.}

\end{tcolorbox}
}
\caption{\textbf{Under-search diagnosed by Belief-Action Gap Diagnosis.} The retrieved evidence identifies Vivian Campbell as the guitarist but does not provide his nationality; despite a negative Belief Score, the agent answers by conflating the band's nationality with the guitarist's personal background.}
\label{case:under_search_gigpo_2}
\end{figure}

\begin{figure}[t]
\centering
\resizebox{\textwidth}{!}{
\begin{tcolorbox}[colback=gray!5!white, colframe=black!75!black, 
title=Prompt Template for Verify-first Action Generation, boxrule=0.3mm, width=\textwidth, arc=3mm, auto outer arc=true]
You are an expert agent tasked with answering the given question step-by-step. Your question: \{task\_description\}. Prior to this step, you have already taken \{step\_count\} step(s). Below is the interaction history where <search> </search> wrapped your past search queries and <information> </information> wrapped the corresponding search results returned by the external search engine. History: \{memory\_context\}. \\

Now it's your turn to respond for the current step. Before making your own decision for this step, a candidate action is proposed as: \{candidate\_decision\}. You should first conduct reasoning, starting by verifying whether the proposed candidate action is appropriate for the current step, then determine the correct action. This process MUST be enclosed within <think> </think> tags. After completing your reasoning, choose only one of the following actions (do not perform both):

(1) If you find you lack some knowledge, you can call a search engine to get more external information using format: <search> your query </search>.

(2) If you have enough knowledge to answer the question confidently, provide your final answer within <answer> </answer> tags, without detailed illustrations. For example, <answer>Beijing</answer>.
\end{tcolorbox}
}
\caption{\textbf{The prompt template for Verify-first Action Generation.}}
\label{prompt:search_verify}
\end{figure}

\begin{figure}[t]
\centering
\resizebox{\textwidth}{!}{
\begin{tcolorbox}[colback=gray!5!white, colframe=black!75!black, 
title=Prompt Template for Internal Belief Probing, boxrule=0.3mm, width=\textwidth, arc=3mm, auto outer arc=true]
You are an expert agent tasked with assessing whether you have enough knowledge to answer the given question. Your question: \{task\_description\}. Prior to this step, you have already taken \{step\_count\} step(s). Below is the interaction history where <search> </search> wrapped your past search queries and <information> </information> wrapped the corresponding search results returned by the external search engine. History: \{memory\_context\}. \\

Respond ONLY with 'Yes' or 'No' to indicate whether you are capable of answering the question confidently:

(1) If you find you lack some knowledge and need to call a search engine to get more external information, output: No.

(2) If you have enough knowledge to answer the question confidently, output: Yes.
\end{tcolorbox}
}
\caption{\textbf{The prompt template for Internal Belief Probing.}}
\label{prompt:search_probe}
\end{figure}

\begin{figure}[t]
\centering
\resizebox{\textwidth}{!}{
\begin{tcolorbox}[colback=gray!5!white, colframe=black!75!black,
title=Prompt Template for External Under-Search Judge, boxrule=0.3mm, width=\textwidth, arc=3mm, auto outer arc=true]
You are an expert evaluator for agentic retrieval-augmented question answering.
Given a question, the retrieved evidence available to the agent, the agent's reasoning before its final answer, the agent's final answer, and the gold answer, determine whether the agent made an \textbf{under-search} error.

An under-search error means that the agent answered prematurely: the available evidence was insufficient, missing a necessary entity or attribute, or was misused in a way that should have triggered another search before answering.

Do \textbf{not} mark under-search if:

(1) the available evidence already supports the correct answer but the agent made a reasoning or extraction mistake;

(2) the final answer is correct;

(3) the error is only due to formatting or wording. \\

Return only one JSON object:
\{
  "under\_search": true or false,
  "reason": "brief explanation"
\}
\end{tcolorbox}
}
\caption{\textbf{The prompt template for External Under-Search Judge.}}
\label{prompt:under_external_judge}
\end{figure}

\begin{figure}[t]
\centering
\resizebox{\textwidth}{!}{
\begin{tcolorbox}[colback=gray!5!white, colframe=black!75!black,
title=Case: Under-Search -- GiGPO (Incorrect), boxrule=0.3mm, width=\textwidth, arc=3mm, auto outer arc=true]

\textbf{Question:} Mold Castle was erected under instructions from the Earl also known by what name? \\
\textbf{Ground Truth:} Hugh the Fat \\[6pt]

\textbf{Step 1}

\textbf{Reasoning:} I need to find out the name by which the Earl who instructed the erection of Mold Castle was also known. I will search the web to find this information.

\textbf{Action:} \texttt{search(``Earl who instructed the erection of Mold Castle also known by what name'')} \\

\textbf{Step 2}

\textbf{Observation:} [Doc 1\&2] Mold Castle \ldots\ is a motte-and-bailey castle erected around 1072, probably by the Norman Robert de Montalt under instructions from \textbf{Hugh d'Avranches, Earl of Chester}. \ldots\ [Doc 3] \ldots\ in the service of \textbf{Hugh Lupus, Earl of Chester}. \ldots

\textbf{Reasoning:} From the information provided, we can see that Mold Castle was erected in 1072 by the Norman Robert de Montalt under the instructions of Hugh d'Avranches, Earl of Chester. The question asks for the name by which the Earl was also known. In the information provided, Hugh d'Avranches is explicitly referred to as the Earl of Chester. Therefore, the answer to the question is the Earl of Chester.

\textbf{Action:} \underline{\texttt{answer(``Earl of Chester'')}} \quad \textit{$\leftarrow$ Under-search: mistakes the title for the alternative name without a follow-up search}

\end{tcolorbox}
}
\caption{\textbf{Under-search failure of GiGPO.} The agent retrieves that Mold Castle was erected under instructions from Hugh d'Avranches, Earl of Chester, but prematurely returns the title ``Earl of Chester'' instead of searching for the requested alias.}
\label{case:under_search_gigpo}
\end{figure}

\begin{figure}[t]
\centering
\resizebox{\textwidth}{!}{
\begin{tcolorbox}[colback=gray!5!white, colframe=black!75!black,
title=Case: Avoiding Under-Search -- \method{} (Correct), boxrule=0.3mm, width=\textwidth, arc=3mm, auto outer arc=true]

\textbf{Question:} Mold Castle was erected under instructions from the Earl also known by what name? \\
\textbf{Ground Truth:} Hugh the Fat \\[6pt]

\textbf{Step 1}

\textbf{Candidate Action Proposed:} \search{}

\textbf{Reasoning:} I lack specific knowledge about Mold Castle and the Earl mentioned in the question. To confidently answer this, I would need to verify the connection between Mold Castle and the Earl's alternative name. Given the uncertainty, I will proceed with a search to find the correct answer.

\textbf{Action:} \texttt{search(``Mold Castle was erected under instructions from the Earl also known by what name'')} \\

\textbf{Step 2}

\textbf{Observation:} [Doc 1\&2] Mold Castle \ldots\ is a motte-and-bailey castle erected around 1072, probably by the Norman Robert de Montalt under instructions from \textbf{Hugh d'Avranches, Earl of Chester}. \ldots\ [Doc 3] \ldots\ in the service of \textbf{Hugh Lupus, Earl of Chester}. \ldots

\textbf{Candidate Action Proposed:} \answer{}

\textbf{Reasoning:} From the search results provided, we can see that Mold Castle was erected under instructions from Hugh d'Avranches, Earl of Chester. To find out the \emph{name by which Hugh d'Avranches was also known}, we can search for this information.

\textbf{Action:} \underline{\texttt{search(``Hugh d'Avranches, Earl of Chester also known as'')}} \quad \textit{$\leftarrow$ Overrides the proposed \answer{} action and searches for the missing alias} \\

\textbf{Step 3}

\textbf{Observation:} [Doc 1\&2] Hugh d'Avranches ($-$ 27 July 1101), also known as \textbf{Hugh the Fat} or \textbf{Hugh Lupus}, was the second Norman Earl of Chester \ldots\ [Doc 3] \ldots\ his father, known as \textbf{Hugh the Fat}, died.

\textbf{Candidate Action Proposed:} \search{}

\textbf{Reasoning:} From the search results provided, we can see that Hugh d'Avranches, Earl of Chester was also known as \textbf{Hugh the Fat}. The question asks for the name by which Hugh d'Avranches was also known, and the answer is given in the search results.

\textbf{Action:} \texttt{answer(``Hugh the Fat'')}

\end{tcolorbox}
}
\caption{\textbf{\method{} avoids under-search on the same question.} The agent recognizes that the first retrieval identifies the Earl but does not yet resolve the requested alias, leading to a targeted follow-up search and the correct answer.}
\label{case:under_search_metarag}
\end{figure}

\end{document}